\pdfoutput=1

\documentclass[sigconf]{acmart}
\AtBeginDocument{%
  }

\setcopyright{acmlicensed}
\copyrightyear{2026}
\acmYear{2026}
\acmDOI{XXXXXXX.XXXXXXX}
\acmConference[KDD '26]{Proceedings of the 32nd ACM SIGKDD Conference on Knowledge Discovery and Data Mining}{August 09--13,
  2026}{Jeju, Korea}
\acmISBN{978-1-4503-XXXX-X/2026/08}

\usepackage{amsmath}

\usepackage{amssymb}
\usepackage{algorithmic}
\usepackage{algorithm}
\usepackage[utf8]{inputenc}
\usepackage{geometry}
\usepackage{enumitem}
\usepackage{booktabs}
\usepackage{multirow}
\usepackage{graphicx}
\usepackage{colortbl}
\usepackage{siunitx} 
\usepackage{makecell}
\usepackage{tabularx}
\usepackage{array}
\usepackage{pgfplots}
\pgfplotsset{compat=1.17}
\usepgfplotslibrary{groupplots}
\usepgfplotslibrary{colormaps}
\usepackage[most]{tcolorbox}
\usepackage{listings}
\usepackage{xcolor}

\usepackage{tcolorbox}
\tcbuselibrary{skins, breakable, listings}
\usepackage{booktabs}
\usepackage{colortbl}
\usepackage{float}
\usepackage{geometry} 

\definecolor{perceptColor}{RGB}{64, 117, 182} 
\definecolor{planColor}{RGB}{209, 156, 95}    
\definecolor{orchColor}{RGB}{165, 86, 73}     
\definecolor{auditColor}{RGB}{117, 156, 81} 

\lstdefinelanguage{json}{
    basicstyle=\ttfamily,
    numbers=none,
    numberstyle=\tiny\color{gray},
    stepnumber=1,
    numbersep=8pt,
    showstringspaces=false,
    breaklines=true,
    frame=none,
    backgroundcolor=\color{white},
    literate=
     *{0}{{{\color{blue}0}}}{1}
      {1}{{{\color{blue}1}}}{1}
      {2}{{{\color{blue}2}}}{1}
      {3}{{{\color{blue}3}}}{1}
      {4}{{{\color{blue}4}}}{1}
      {5}{{{\color{blue}5}}}{1}
      {6}{{{\color{blue}6}}}{1}
      {7}{{{\color{blue}7}}}{1}
      {8}{{{\color{blue}8}}}{1}
      {9}{{{\color{blue}9}}}{1}
      {:}{{{\color{red}{:}}}}{1}
      {,}{{{\color{red}{,}}}}{1}
      {\{}{{{\color{black}{\{}}}}{1}
      {\}}{{{\color{black}{\}}}}}{1}
      {[}{{{\color{black}{[}}}}{1}
      {]}{{{\color{black}{]}}}}{1},
}

\newtcolorbox{agentbox}[2][]{
    enhanced,
    breakable,
    colback=white,
    colframe=#2,
    coltitle=white,
    title=\textbf{#1},
    fonttitle=\sffamily,
    boxrule=0.5mm,
    arc=2mm,
    left=2mm, right=2mm, top=2mm, bottom=2mm,
    drop shadow
}

\definecolor{dynamicblue}{HTML}{0073E6} 
\definecolor{fallgreen}{HTML}{008000} 
\definecolor{risered}{HTML}{CC0000}   
\newcommand{\grad}[3]{\cellcolor{dynamicblue!#1}#2~#3}

\newcommand{\rise}[1]{\textcolor{risered}{\scriptsize{(+#1)}}}
\newcommand{\fall}[1]{\textcolor{fallgreen}{\scriptsize{(#1)}}}

\begin{document}

\title{VentAgent: When LLMs Learn to Breathe: Multi-Objective Arbitration for ARDS Ventilation}

\author{Teqi Hao}
\authornote{Both authors contributed equally to this work.}
\affiliation{%
  \institution{School of Electronic and Electrical Engineering, Shanghai University of Engineering Science}
  \city{Shanghai}
  \state{Shanghai}
  \country{China}
}

\author{Yuxuan Fu}
\authornotemark[1]
\affiliation{%
  \institution{School of Electronic and Electrical Engineering, Shanghai University of Engineering Science}
  \city{Shanghai}
  \state{Shanghai}
  \country{China}
}

\author{Xiaoyu Tan}
\authornotemark[1]
\affiliation{%
  \institution{Tencent Youtu Lab}
  \city{Shanghai}
  \state{Shanghai}
  \country{China}
}

\author{Shaojie Shi}
\affiliation{%
  \institution{Artificial Intelligence Innovation and Incubation Institute, Fudan University}
  \city{Shanghai}
  \state{Shanghai}
  \country{China}
}

\author{Bohao Lv}
\affiliation{%
  \institution{Artificial Intelligence Innovation and Incubation Institute, Fudan University}
  \city{Shanghai}
  \state{Shanghai}
  \country{China}
}

\author{Yinghui Xu}
\affiliation{%
  \institution{Artificial Intelligence Innovation and Incubation Institute, Fudan University}
  \city{Shanghai}
  \state{Shanghai}
  \country{China}
}

\author{Xihe Qiu}
\authornote{Corresponding author.}
\affiliation{%
  \institution{School of Electronic and Electrical Engineering, Shanghai University of Engineering Science}
  \city{Shanghai}
  \state{Shanghai}
  \country{China}
}
\email{qiuxihe1993@gmail.com}

\renewcommand{\shortauthors}{Hao et al.}

\begin{abstract}
Mechanical ventilation for Acute Respiratory Distress Syndrome (ARDS) mandates a delicate equilibrium between competing physiological objectives: oxygenation, lung protection, and acid-base homeostasis. However, current data-driven paradigms, particularly those imitating retrospective Electronic Health Records (EHR), suffer fundamentally from \textit{imitation bias}. Much like an image classifier relying on background scenery rather than the object, these models often capture superficial statistical correlations from inconsistent human demonstrations (e.g., associating passive settings with survival simply because they are prevalent in stable patients), failing to generalize to volatile, out-of-distribution phenotypes. Furthermore, standard Reinforcement Learning (RL) methods struggle to navigate the adversarial trade-offs inherent in critical care, often collapsing into opaque ``black boxes'' that lack clinical safety assurances. To address these structural deficiencies, we introduce \textbf{VentAgent}, a hierarchical framework where Large Language Models (LLMs) learn to ``breathe'' by acting as transparent arbitrators. We reformulate the ventilation task not as a single-objective optimization, but as a dynamic \textit{Multi-Objective Arbitration} process. \textsc{VentAgent} decouples decision-making into three interpretable stages: \textit{Perception}, \textit{Planning}, and \textit{Orchestration}. By leveraging the semantic reasoning capabilities of LLMs, the system synthesizes diverse strategies from heterogeneous experts and resolves conflicting clinical priorities through an explicit coordination mechanism. Systematic evaluations on a high-fidelity physiological simulator demonstrate that \textsc{VentAgent} significantly outperforms state-of-the-art RL and classical control baselines. Crucially, it transforms the control policy into a human-readable reasoning chain, establishing a new paradigm for safe, interpretable, and adaptable critical care automation.
\end{abstract}




\maketitle

\section{Introduction}
\begin{figure*}[t]
    \centering
    \includegraphics[width=\textwidth]{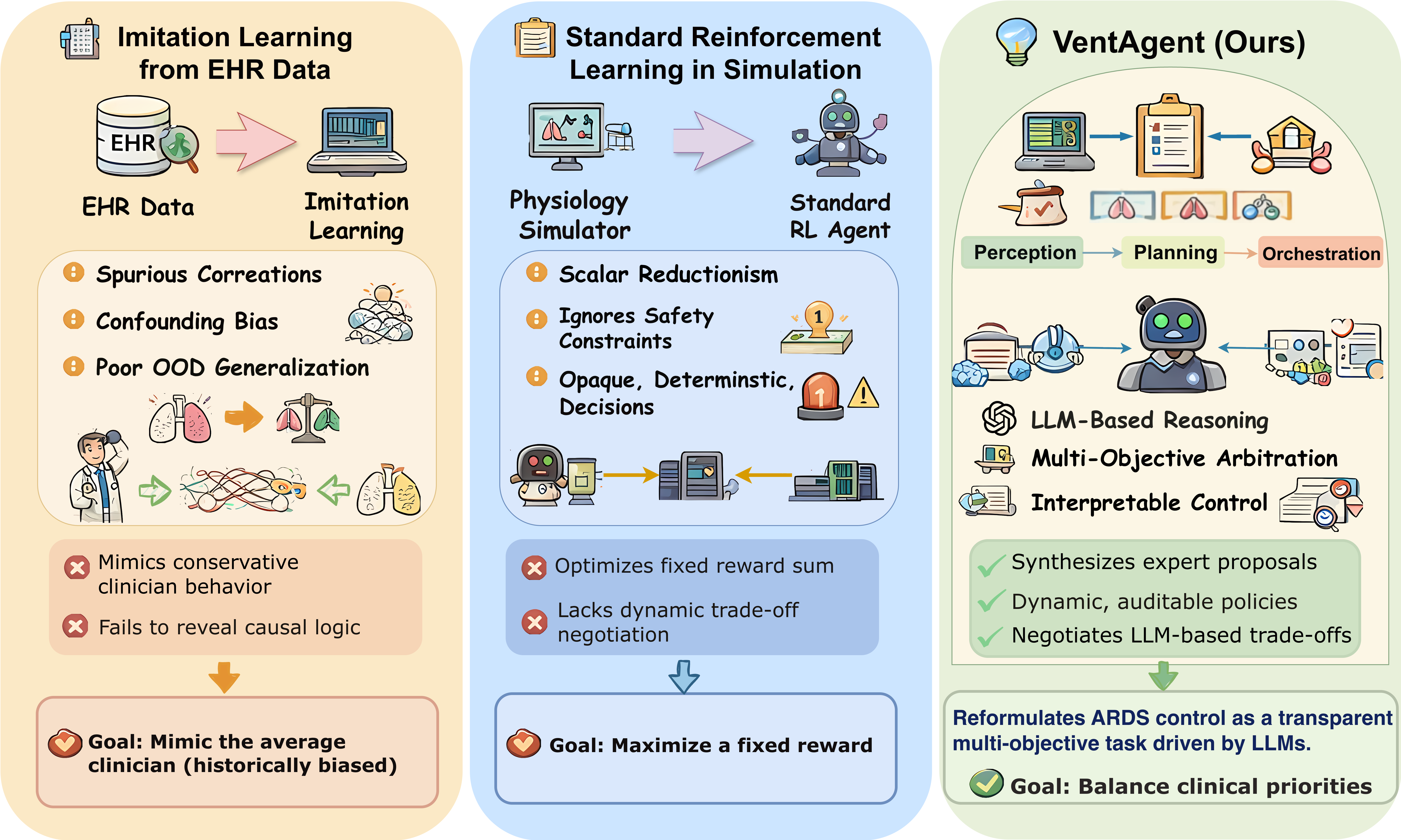}
    \caption{Comparison of three paradigms for ARDS ventilation control. Imitation Learning from EHRs suffers from spurious correlations and fails to uncover causal logic. Standard Reinforcement Learning collapses multi-objective goals into a scalar reward, ignoring safety constraints. Our proposed \textsc{VentAgent} reformulates the problem as interpretable, multi-objective control via LLM-based reasoning, enabling transparent arbitration of conflicting clinical priorities.}
    \label{fig:paradigm_comparison}
\end{figure*}

The clinical management of Acute Respiratory Distress Syndrome (ARDS) represents a quintessential high-stakes control challenge within critical care medicine \citep{ranieri2012acute}. Characterized by profound physiological heterogeneity, ranging from sepsis-induced alveolar collapse to trauma-related edema, ARDS necessitates a ventilation strategy that maintains a delicate dynamic equilibrium among competing therapeutic objectives. The crux of this challenge lies in navigating \textit{adversarial trade-offs}: aggressive oxygenation often necessitates high pressures that risk barotrauma (lung injury), while lung-protective strategies may lead to permissive hypercapnia (acidosis) \citep{acute2000ventilation}. Crucially, these trade-offs are highly phenotype-specific; a policy that is optimal for a patient with high lung recruitability may prove catastrophic for one with high dead-space ventilation \citep{calfee2014subphenotypes}. This complexity imposes an immense cognitive load on clinicians, demanding high-frequency, meticulous adjustments that often exceed human bandwidth in resource-constrained Intensive Care Units (ICUs).

While data-driven automation offers a promising solution, existing methodologies suffer from systemic structural deficiencies that limit their clinical deployment.
\textbf{First, regarding Imitation Learning (IL) from Electronic Health Records (EHRs):} These approaches are fundamentally constrained by \textit{spurious correlations} and \textit{confounding biases}. EHR data records actions taken by clinicians, but rarely the unobserved counterfactual reasoning behind them \citep{wornow2023shaky}. For instance, aggressive ventilation might statistically correlate with higher mortality not because it is harmful, but because it is typically reserved for the sickest patients. Models trained on such data learn to mimic the ``average'' clinician behavior, often conservative and sub-optimal, rather than discovering the causal decision logic required for precise physiological regulation \citep{komorowski2018artificial, gottesman2019guidelines, levine2020offline,richens2020improving}. They fail to capture the ``why'' behind a trade-off, leading to poor generalization on out-of-distribution (OOD) phenotypes.

\textbf{Second, regarding Reinforcement Learning (RL) in simulation:} To overcome data sparsity and static biases, recent research has pivoted towards high-fidelity simulators (e.g., the Pulse Physiology Engine) \citep{bray2019pulse}. However, standard RL paradigms typically collapse the rich, conflicting spectrum of clinical goals into a single scalar reward function \citep{raghu2017deep}. We contend that this \textit{scalar reductionism} is fundamentally flawed for critical care. Optimal ARDS management is not a maximization problem of a single numerical value; it is a \textit{dynamic multi-objective arbitration} task. By forcing an agent to optimize a fixed weighted sum, existing RL methods obscure the inherent safety constraints and produce opaque, deterministic policies that lack the flexibility to negotiate trade-offs dynamically (e.g., explicitly sacrificing partial oxygenation to prevent imminent lung rupture).

To address these limitations, we propose \textsc{VentAgent}, a hierarchical framework that reformulates ventilation control as a transparent arbitration process driven by Large Language Models (LLMs) \citep{xi2025rise}. Our core insight is that LLMs, unlike numerical optimizers, possess the semantic reasoning capabilities to explicitize and negotiate clinical trade-offs \citep{singhal2023large, nori2023capabilities}. Rather than blindly maximizing a reward, \textsc{VentAgent} learns to ``breathe'' by emulating the cognitive process of a senior clinician: observing physiological states (\textit{Perception}), soliciting proposals from diverse functional experts (\textit{Planning}), and arbitrating the final decision based on safety boundaries and patient-specific needs (\textit{Orchestration}) \citep{wei2022chain, yao2022react}.

This work establishes a new paradigm for interpretable critical care automation. Our main contributions are:
\begin{itemize}
    \item \textbf{Problem Reformulation:} We redefine ARDS ventilation not as scalar reward maximization, but as a constrained multi-objective arbitration problem, identifying the structural failures of scalar RL in adversarial medical settings.
    \item \textbf{The VentAgent Framework:} We introduce a novel three-stage architecture (Perception-Planning-Orchestration) that leverages heterogeneous LLM-based experts to synthesize conflicting clinical priorities into a unified, safe strategy.
    \item \textbf{Interpretability by Design:} Unlike ``black-box'' neural networks, \textsc{VentAgent} generates structured, human-readable reasoning logs at every decision step, ensuring the decision trajectory is fully auditable and clinically transparent.
    \item \textbf{Empirical Validation:} We conduct systematic evaluations on a high-fidelity physiological simulator across twenty distinct patient phenotypes. Results demonstrate that \textsc{VentAgent} significantly outperforms state-of-the-art RL and classical PID baselines in both safety and stability.
\end{itemize}

\section{Related Work}
\subsection{Data-Driven Decision Support in Critical Care}
Recent advances in retrospective medical AI have predominantly focused on leveraging Electronic Health Records (EHRs) for outcome prediction and treatment recommendation \citep{johnson2016mimic, johnson2023mimic, wornow2023shaky}. Reinforcement Learning (RL) approaches, such as Deep Q-Networks (DQN) and Inverse RL, have been applied to optimize sepsis management and mechanical ventilation \citep{komorowski2018artificial, prasad2017reinforcement, yu2021reinforcement, kondrup2023towards}. However, these methods face the intrinsic \textit{"Deadly Triad"} of offline medical RL: data sparsity, irregular sampling, and, most critically, \textit{confounding bias} \cite{gottesman2019guidelines, futoma2020popcorn}. Since EHRs only record executed actions without their counterfactual alternatives, RL agents often learn to mimic conservative or biased clinician behaviors rather than optimal physiological regulation \citep{oberst2019counterfactual, jiang2016doubly, thomas2016data}. Furthermore, deep RL models typically operate as opaque "black boxes," obscuring the rationale behind life-critical decisions. In contrast, \textsc{VentAgent} eschews purely statistical correlation in favor of LLM-driven causal reasoning, integrating medical domain knowledge to ensure clinical consistency and interpretability.

\subsection{Simulation-Based Control \& The Scalar Reward Trap}
To circumvent the safety risks of exploring policies on real patients, high-fidelity physiological simulators (e.g., Pulse Engine, SimGlucose) have emerged as essential testbeds for closed-loop control \cite{bray2019pulse}. While simulation enables active exploration, existing RL-based control frameworks remain structurally limited by their objective formulation \citep{roijers2013survey, levine2020offline}. Predominant methods collapse complex, conflicting clinical goals (e.g., minimizing driving pressure vs. maintaining permissive hypercapnia) into a single scalar reward function \cite{ward2024optimal, roijers2013survey}. This \textit{scalar reductionism} forces the agent to learn a rigid policy that maximizes a numerical value, often failing to navigate the dynamic Pareto frontier of patient safety \citep{garcia2015comprehensive,altman2021constrained}. Moreover, many approaches bootstrap from fixed clinical guidelines, restricting the agent's ability to adapt to severe, out-of-distribution phenotypes \citep{ranieri2012acute,acute2000ventilation, amato2015driving}. \textsc{VentAgent} departs from this paradigm by reformulating control as a \textit{multi-objective arbitration} task, using LLMs to dynamically negotiate trade-offs rather than blindly optimizing a fixed reward sum \citep{achiam2017constrained}.

\subsection{LLM Agents in Medicine: From QA to Active Control}
Large Language Models (LLMs) have demonstrated impressive capabilities in medical Question Answering (QA) and diagnostic reasoning (e.g., Med-PaLM) \citep{tu2024towards, singhal2023large, nori2023capabilities, singhal2025toward}. However, the transition from "passive consultant" to "active agent" remains nascent \citep{shen2023hugginggpt, xi2025rise}. Most current medical agents operate in \textit{open-loop} settings (e.g., report generation, consultation), detached from real-time physiological feedback loops \citep{tang2024medagents}. While recent works explore agents for software tasks or embodied robotics, their application in high-stakes physiological control is hampered by hallucinations and the lack of precise temporal regulation. Furthermore, existing architectures are often monolithic, struggling to handle the multidisciplinary complexity of ICU care \citep{ji2023survey}. \textsc{VentAgent} addresses these gaps by introducing a hierarchical, multi-expert agent framework specifically designed for \textit{closed-loop} critical care, bridging the gap between semantic medical reasoning and precise continuous control.

\section{Preliminaries}

\noindent\textbf{Problem Formulation.} \
We formalize ARDS ventilation management as a \textit{Constrained Markov Decision Process} (CMDP) \citep{achiam2017constrained,altman2021constrained}, defined by the tuple $\mathcal{M} = \langle \mathcal{S}, \mathcal{A}, \mathcal{P}, \mathbf{R}, \mathcal{C}, \gamma \rangle$. Unlike standard MDPs with explicit transition matrices, the system dynamics $\mathcal{P}: \mathcal{S} \times \mathcal{A} \to \mathcal{S}$ are governed by the high-fidelity Pulse Physiology Engine, acting as an implicit environment \citep{bray2019pulse}. At each discrete time step $t$, the agent observes a physiological state $\mathbf{s}_t \in \mathcal{S}$ and executes a continuous control action $\mathbf{a}_t \in \mathcal{A}$. The objective is to maximize a vector-valued clinical utility $\mathbf{R}(\mathbf{s}_t, \mathbf{a}_t)$ subject to safety constraints $\mathcal{C}$.

\vspace{0.5em}
\noindent\textbf{State and Action Spaces.} \
The observation space $\mathcal{S} \subseteq \mathbb{R}^{24}$ represents a comprehensive physiological snapshot, structured as the concatenation ($\oplus$) of four clinical sub-domains: $\mathbf{s}_t = [\mathbf{x}_{\text{oxy}} \oplus \mathbf{x}_{\text{vent}} \oplus \mathbf{x}_{\text{mech}} \oplus \mathbf{x}_{\text{hemo}}]^\top$, corresponding to gas exchange, acid-base balance, lung mechanics, and hemodynamic stability, respectively.
The action space $\mathcal{A} \subseteq \mathbb{R}^6$ controls the Volume Control Continuous Mandatory Ventilation (VC-CMV) mode \citep{fan2017official}. The action vector is defined as $\mathbf{a}_t = [\text{PEEP}, \text{FiO}_2, \text{RR}, V_T, T_{\text{insp}}, \dot{V}]^\top$, bounded by clinical safety limits.

\vspace{0.5em}
\noindent\textbf{Clinical Utility Functions.} \
To map complex physiological states to scalar rewards, we define three generic shaping primitives: a \textit{Target Retention} sigmoid $\Phi$, a \textit{Minimization} decay $\mathcal{L}$, and a \textit{Hard Constraint} penalty $\Theta$ \citep{ng1999policy}. We formulate these as:
\begin{align}
    \Phi(x; x^*, \epsilon, k) &= \frac{2}{1 + \exp(k(|x - x^*| - \epsilon))} - 1, \\
    \mathcal{L}(x; \tau) &= 1 - \lambda_{\text{dec}} [x - \tau]^+, \\
    \Theta(x; \tau) &= -\lambda_{\text{step}} [x - \tau]^+,
\end{align}
where $[z]^+ \triangleq \max(0, z)$. Based on these primitives, the vector-valued reward $\mathbf{r}_t = [r_{\text{oxy}}, r_{\text{vent}}, r_{\text{mech}}]^\top$ is constructed as follows:

\vspace{0.5em}
\noindent\textbf{(1) Oxygenation ($r_{\text{oxy}}$):} \
We employ a dynamic target $\text{SpO}_2^*$ modulated by patient severity (via the P/F ratio coefficient $\alpha_t$). The reward penalizes deviations from this target while strictly limiting oxygen toxicity:
\begin{equation}
    r_{\text{oxy}} = \mathcal{U}_{\text{sat}}(\text{SpO}_2; \text{SpO}_2^*) - w_{\text{tox}} [\text{FiO}_2 - \xi_{\text{tox}}]^{+2}.
\end{equation}

\noindent\textbf{(2) Ventilation ($r_{\text{vent}}$):} \
Prioritizing acid-base homeostasis, we introduce a \textit{conditional gating mechanism}. $\text{PaCO}_2$ is rewarded only if pH is within a safe range ($\text{pH} > \tau_{\text{gate}}$), otherwise purely pH optimization takes precedence:
\begin{equation}
    r_{\text{vent}} = w_{\text{pH}} \Phi(\text{pH}) + w_{\text{CO}_2} \mathbb{I}(\text{pH} > \tau_{\text{gate}}) r_{\text{CO}_2} + w_{\text{RR}} \Phi(\text{RR}).
\end{equation}

\noindent\textbf{(3) Mechanics \& Safety ($r_{\text{mech}}$):} \
This term enforces lung protection by penalizing mechanical power ($P_{\text{pow}}$), driving pressure ($\Delta P$), and plateau pressure ($P_{\text{plat}}$) \citep{gattinoni2016ventilator, slutsky2013ventilator, amato2015driving, fan2017official}. A critical hemodynamic override is applied to prevent cardiovascular collapse:
\begin{equation}
    r_{\text{mech}} = 
    \begin{cases} 
    -1, & \text{if } \text{MAP} < \tau_{\text{MAP}} \land \text{PEEP} > \tau_{\text{PEEP}} \\
    r_{\text{safe}}, & \text{otherwise},
    \end{cases}
\end{equation}
where the stable state reward is defined as $r_{\text{safe}} =  \\ \sum_{k \in \{\Delta P, \text{pow}\}} w_k \mathcal{L}(k; \tau_k) + w_{\text{plat}}\Theta(P_{\text{plat}})$.

\section{Methodology}
\begin{figure*}[t]
  \centering
  \includegraphics[width=\linewidth]{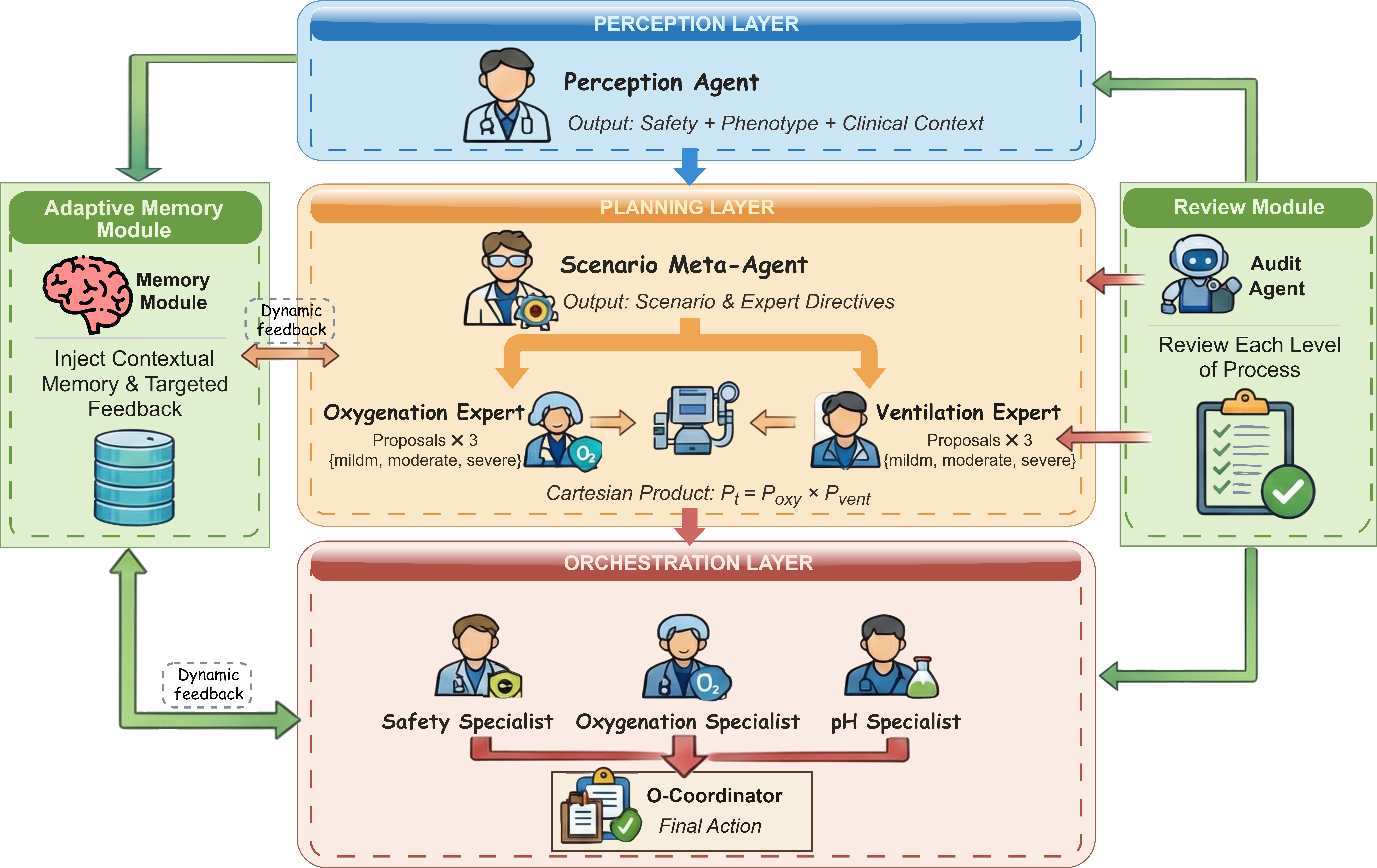} 
  \caption{\textbf{The Hierarchical Architecture of VentAgent.} Grounded in the principle of \textit{Cognitive Decoupling}, the framework decomposes the decision process into three logical stages: (1) \textbf{Perception Layer} projects raw telemetry into a semantic manifold $\hat{\mathbf{s}}_t$; (2) \textbf{Planning Layer} generates diverse candidate strategies $\mathcal{P}_t$ guided by meta-scenarios; (3) \textbf{Orchestration Layer} resolves adversarial trade-offs via multi-perspective arbitration. Parallel \textit{Memory} and \textit{Audit} modules ensure adaptive evolution and clinical safety boundaries.}
  \label{fig:architecture}
\end{figure*}

We propose \textbf{VentAgent}, a hierarchical multi-agent decision framework designed to emulate the cognitive workflow of an ICU multidisciplinary team. Grounded in the principle of \textit{cognitive decoupling}, the framework addresses the intractability of ARDS management by decomposing the clinical decision process into three progressive reasoning stages: Perception, Planning, and Orchestration.

As illustrated in Fig.~\ref{fig:architecture}, at each timestep $t$, the system transforms raw high-frequency telemetry into interpretable diagnostic semantics, navigates a complex action space through dynamically constructed experts, and achieves cross-dimensional value alignment via multi-perspective arbitration. A sparse reflective memory module further facilitates closed-loop evolution by refining policies based on historical trajectories.

\subsection{LLM as a Sequential Decision Operator}
Unlike traditional RL approaches that parameterize policy as a ``black-box'' neural network $\pi_\theta(\mathbf{a}|\mathbf{s})$, we conceptualize the Large Language Model (LLM) as a \textit{Clinical Reasoning Operator} augmented with domain logic. Formally, let $\mathcal{L}$ denote the pre-trained LLM. We define a generic reasoning agent $\mathcal{G}_i$ as a mapping function:
\begin{equation}
    \mathbf{o}_i = \mathcal{G}_i(\mathbf{s}_t, \mathcal{K}_i \mid \mathcal{P}_i) \xrightarrow{\mathcal{L}} \text{Reasoning Chain}
    \label{eq:agent}
\end{equation}
where $\mathbf{s}_t$ is the physiological state, $\mathcal{K}_i$ represents injected domain knowledge (e.g., ARDSNet protocols), and $\mathcal{P}_i$ is the role-specific system prompt. The output $\mathbf{o}_i$ is a structured decision artifact derived via in-context learning \citep{brown2020language}. The global decision process is modeled as a compositional function chain:
\begin{equation}
    \mathbf{a}_t^* = \underbrace{\mathcal{G}_{\text{coord}} \circ \mathcal{G}_{\text{plan}}}_{\text{Decision Synthesis}} \circ \underbrace{\mathcal{G}_{\text{percept}}}_{\text{State Abstraction}}(\mathbf{s}_t)
    \label{eq:composition}
\end{equation}
This architecture ensures that decision-making is not merely probabilistic token generation, but a structured derivation based on explicit clinical logic.

\subsection{Perception Layer: Semantic State Abstraction}
To bridge the gap between high-frequency telemetry and clinical cognition, we construct the \textit{Perception Layer} as a unified diagnostic agent ($\mathcal{G}_{\text{percept}}$). Instead of processing raw signals in isolation, this agent projects the high-dimensional non-linear state $\mathbf{s}_t$ into a latent semantic manifold $\hat{\mathbf{s}}_t$:
\begin{equation}
    \hat{\mathbf{s}}_t = \mathcal{G}_{\text{percept}}(\mathbf{s}_t) = \langle \mathcal{H}_{\text{sev}}, \phi_{\text{pheno}}, \rho_{\text{recruit}}, \mathcal{C}_{\text{active}} \rangle
\end{equation}
Specifically, it performs joint reasoning to extract: (1) \textbf{Disease Severity} $\mathcal{H}_{\text{sev}}$ and \textbf{Inflammatory Phenotype} $\phi_{\text{pheno}}$; (2) \textbf{Recruitability Potential} $\rho_{\text{recruit}}$; and crucially, (3) \textbf{Active Constraints} $\mathcal{C}_{\text{active}}$ (e.g., "Driving Pressure limit $< 14$ cmH$_2$O"). This semantic abstraction $\hat{\mathbf{s}}_t$ serves as the immutable ground truth for all downstream modules, ensuring decisions are grounded in a holistic pathological understanding rather than transient numerical fluctuations.

\subsection{Planning Layer: Meta-Scenario Synthesis and Divergent Proposal}
The planning phase addresses the combinatorial explosion of the action space. We introduce a \textbf{Scenario Meta-Agent} ($\mathcal{G}_{\text{meta}}$) to first synthesize the semantic state $\hat{\mathbf{s}}_t$ into a high-level strategic directive $\mathcal{S}_t$ (e.g., "Maximal Recruitment" vs. "Weaning Preparation"). This directive effectively prunes the search space, constraining downstream experts to physiologically relevant strategies.

Guided by $\mathcal{S}_t$, two domain-specific experts operate in parallel to generate a candidate action set $\mathcal{P}_t$:
\begin{itemize}
    \item \textbf{Oxygenation Expert ($\mathcal{G}_{\text{oxy}}$):} Focuses on gas exchange (PEEP, FiO$_2$), balancing alveolar recruitment benefits against hemodynamic risks.
    \item \textbf{Ventilation Expert ($\mathcal{G}_{\text{vent}}$):} Manages acid-base homeostasis (RR, V$_T$, flow), aiming to minimize mechanical power.
\end{itemize}
Crucially, each expert generates a \textit{gradient of proposals} (Conservative, Standard, Aggressive). The final candidate pool $\mathcal{P}_t = \mathcal{P}_{\text{oxy}} \times \mathcal{P}_{\text{vent}}$ represents the Cartesian product of these domain-specific trajectories, providing a structured exploration space for the subsequent arbitration.

\subsection{Orchestration Layer: Multi-Perspective Arbitration}
The Orchestration Layer resolves the inherent adversarial trade-offs (e.g., oxygenation vs. lung protection). We employ a generative arbitration mechanism where three \textbf{Perspective Refinement Agents} acting as specialized consultants—\textit{Pulmonary Protection} ($\mathcal{G}_{\text{protect}}$), \textit{Oxygenation Security} ($\mathcal{G}_{\text{secure}}$), and \textit{Acid-Base Homeostasis} ($\mathcal{G}_{\text{base}}$)—critique the candidate plans against their specific utility functions.

The \textbf{Clinical Coordinator} ($\mathcal{G}_{\text{coord}}$) synthesizes the final strategy $\mathbf{a}_t^*$ by aggregating these conflicting perspectives:
\begin{equation}
    \mathbf{a}_t^* = \underset{\mathbf{a} \in \mathcal{P}_t}{\text{arbitrate}} \left( \{\mathbf{a}_{\text{protect}}, \mathbf{a}_{\text{secure}}, \mathbf{a}_{\text{base}}\} \mid \hat{\mathbf{s}}_t, \mathcal{C}_{\text{active}} \right)
\end{equation}
Unlike simple weighted averaging, $\mathcal{G}_{\text{coord}}$ applies hierarchical logic to negotiate deadlocks (e.g., prioritizing mechanical safety limits over pH correction during "stiff lung" episodes), ensuring the final action lies on the optimal Pareto frontier of patient safety.


\subsection{Sparse Reflective Memory: Closed-Loop Evolution}
To mitigate the limitations of isolated single-step reasoning, we introduce a \textbf{Memory Agent} ($\mathcal{G}_{\text{mem}}$) that functions as a meta-cognitive observer. At each timestep, $\mathcal{G}_{\text{mem}}$ evaluates the alignment between the predicted outcome and the actual environmental feedback $\mathbf{r}_t$.
Successful reasoning traces and identified attribution errors are distilled into semantic reflections $\mathcal{M}_t$. These reflections are injected into the context of subsequent decision cycles as targeted feedback signals, enabling the system to perform online policy refinement (e.g., "Patient exhibits CO$_2$ retention tendency, increase baseline RR") without parameter updates \citep{shinn2023reflexion, madaan2023self}.

\begin{table*}[t]
    \centering
    \renewcommand{\arraystretch}{1.2}
    \setlength{\tabcolsep}{1.5pt}
    \caption{Performance comparison of LLM-based methods. The proposed VentAgent demonstrates superior robustness. Values in parentheses indicate the performance gap ($\downarrow$) of baselines relative to VentAgent.}
    \label{tab:llm_results}
    \resizebox{\textwidth}{!}{%
    \begin{tabular}{lccccccccc}
    \toprule
    \multirow{2.5}{*}{\textbf{Method}} & \multicolumn{3}{c}{\textbf{Mild}} & \multicolumn{3}{c}{\textbf{Moderate}} & \multicolumn{3}{c}{\textbf{Severe}} \\
    \cmidrule(lr){2-4} \cmidrule(lr){5-7} \cmidrule(lr){8-10}
     & \textbf{$r_{oxy}$} & \textbf{$r_{vent}$} & \textbf{$r_{mech}$} & \textbf{$r_{oxy}$} & \textbf{$r_{vent}$} & \textbf{$r_{mech}$} & \textbf{$r_{oxy}$} & \textbf{$r_{vent}$} & \textbf{$r_{mech}$} \\
    \midrule
    Few-Shot & \grad{15}{2155.17}{\fall{-725.24}} & \grad{10}{160.14}{\fall{-178.13}} & \grad{10}{548.38}{\fall{-477.16}} & \grad{15}{1945.63}{\fall{-750.26}} & \grad{10}{62.19}{\fall{-97.93}} & \grad{10}{305.82}{\fall{-282.64}} & \grad{15}{1488.45}{\fall{-751.88}} & \grad{5}{-142.56}{\fall{-99.98}} & \grad{10}{82.11}{\fall{-93.53}} \\
    CoT & \grad{18}{2265.42}{\fall{-614.99}} & \grad{12}{188.29}{\fall{-149.98}} & \grad{12}{598.14}{\fall{-427.40}} & \grad{18}{2110.37}{\fall{-585.52}} & \grad{12}{78.43}{\fall{-81.69}} & \grad{12}{345.58}{\fall{-242.88}} & \grad{18}{1605.81}{\fall{-634.52}} & \grad{8}{-128.24}{\fall{-85.66}} & \grad{12}{95.67}{\fall{-79.97}} \\
    \addlinespace[4pt]
    ReAct & \grad{22}{2490.86}{\fall{-389.55}} & \grad{18}{212.53}{\fall{-125.74}} & \grad{18}{668.49}{\fall{-357.05}} & \grad{22}{2265.12}{\fall{-430.77}} & \grad{15}{92.68}{\fall{-67.44}} & \grad{18}{378.34}{\fall{-210.12}} & \grad{22}{1745.27}{\fall{-495.06}} & \grad{12}{-110.85}{\fall{-68.27}} & \grad{15}{108.41}{\fall{-67.23}} \\
    Reflexion & \grad{25}{2825.64}{\fall{-54.77}} & \grad{20}{248.18}{\fall{-90.09}} & \grad{20}{770.21}{\fall{-255.33}} & \grad{25}{2580.46}{\fall{-115.43}} & \grad{18}{125.59}{\fall{-34.53}} & \grad{20}{435.13}{\fall{-153.33}} & \grad{25}{1995.62}{\fall{-244.71}} & \grad{15}{-102.37}{\fall{-59.79}} & \grad{18}{132.55}{\fall{-43.09}} \\
    \addlinespace[4pt]
    Self-Consist. & \grad{20}{2340.28}{\fall{-540.13}} & \grad{15}{195.84}{\fall{-142.43}} & \grad{15}{620.41}{\fall{-405.13}} & \grad{20}{2175.56}{\fall{-520.33}} & \grad{12}{85.39}{\fall{-74.73}} & \grad{15}{360.22}{\fall{-228.24}} & \grad{20}{1698.47}{\fall{-541.86}} & \grad{10}{-118.63}{\fall{-76.05}} & \grad{12}{105.88}{\fall{-69.76}} \\
    Debate & \grad{22}{2575.13}{\fall{-305.28}} & \grad{18}{218.49}{\fall{-119.78}} & \grad{18}{690.12}{\fall{-335.42}} & \grad{22}{2360.85}{\fall{-335.04}} & \grad{15}{98.24}{\fall{-61.88}} & \grad{18}{395.57}{\fall{-192.89}} & \grad{22}{1850.31}{\fall{-390.02}} & \grad{12}{-105.46}{\fall{-62.88}} & \grad{15}{115.23}{\fall{-60.41}} \\
    ToT & \grad{24}{2692.59}{\fall{-187.82}} & \grad{18}{232.65}{\fall{-105.62}} & \grad{18}{728.34}{\fall{-297.20}} & \grad{24}{2485.27}{\fall{-210.62}} & \grad{15}{112.43}{\fall{-47.69}} & \grad{18}{415.89}{\fall{-172.57}} & \grad{24}{1955.15}{\fall{-285.18}} & \grad{12}{-95.82}{\fall{-53.24}} & \grad{15}{128.46}{\fall{-47.18}} \\
    \midrule
    \textbf{VentAgent} & \grad{40}{\textbf{2880.41}}{} & \grad{35}{\textbf{338.27}}{} & \grad{35}{\textbf{1025.54}}{} & \grad{40}{\textbf{2695.89}}{} & \grad{30}{\textbf{160.12}}{} & \grad{30}{\textbf{588.46}}{} & \grad{40}{\textbf{2240.33}}{} & \grad{25}{\textbf{-42.58}}{} & \grad{30}{\textbf{175.64}}{} \\
    \bottomrule
    \end{tabular}%
    }
\end{table*}

\subsection{Automated Trajectory Curation via Hierarchical Audit}
To safeguard clinical integrity under LLM stochasticity, we deploy a parallel \textbf{Audit Agent} ($\mathcal{G}_{\text{audit}}$) that performs layer-wise verification. We formalize this as a verification function $\Psi$:
\begin{equation}
    v_{k}, \tilde{\mathbf{o}}_{k} = \Psi(\mathbf{o}_{k} \mid \mathbf{s}_t, \mathcal{K}_{\text{med}}, \mathcal{C}_{\text{consistency}})
\end{equation}
where $\mathbf{o}_{k}$ is the intermediate output from any layer $k$. The Audit Agent verifies three criteria: (1) \textbf{Factual Accuracy} against physiological laws; (2) \textbf{Logical Coherence} between diagnosis $\hat{\mathbf{s}}_t$ and action; and (3) \textbf{Completeness} of safety constraints. If a violation is detected ($v_k=0$), the module triggers a regeneration or replaces the output with a rectified trace $\tilde{\mathbf{o}}_{k}$. This "Cognitive Firewall" ensures that the final decision trajectory remains strictly within the bounds of clinical safety protocols \citep{bai2022constitutional, madaan2023self}.

\section{Experiments}

\subsection{Experimental Setups}

\noindent\textbf{Benchmark and Patient Cohorts.} \
The efficacy of VentAgent is assessed using the Pulse Physiology Engine, a high-fidelity simulation platform that employs equivalent circuit modeling to represent complex respiratory mechanics. The treatment cycle spans 24 hours, discretized into $T=48$ steps with an intervention interval of $\Delta t = 30$ minutes. 
To ensure robust evaluation, we define two distinct cohorts: 
(1) The \textit{Sampling Cohort}, comprising 20 unique patient configurations across three ARDS severities (mild, moderate, and severe), used to generate interaction data totaling $2,880$ state-action-state triplets; 
(2) The \textit{Evaluation Cohort}, consisting of 100 virtual patients with a balanced demographic split (50 males, 50 females; ages 18--65), held out to rigorously assess generalization performance on unseen phenotypes.

\noindent\textbf{Baselines.} \
We comprehensively evaluate our framework against three classes of LLM inference paradigms, all utilizing \texttt{gpt-4o-mini} \citep{menick2024gpt} as the backbone model to ensure fair comparison:
\begin{itemize}
    \item \textbf{Linear Reasoning:} \textit{Few-shot} (providing clinical exemplars) and \textit{CoT} (Chain-of-Thought).
    \item \textbf{Feedback-Driven Reasoning:} \textit{ReAct} \citep{yao2022react} (interleaving reasoning and action generation) and \textit{Reflexion} \citep{shinn2023reflexion} (using verbal reinforcement to critique past actions).
    \item \textbf{Ensemble Reasoning:} \textit{Tree of Thoughts (ToT)} \citep{yao2023tree} (exploring multiple reasoning branches via BFS/DFS), \textit{Debate} \citep{du2023improving} (multi-agent discussion without hierarchical arbitration), and \textit{Self-Consistency} \citep{wang2022self} (majority voting on diverse reasoning paths).
\end{itemize}

\subsection{Main Results}
\label{sec:main}

The quantitative evaluation, detailed in Table~\ref{tab:llm_results}, substantiates a significant performance disparity between VentAgent and existing baselines. 
While linear and feedback-driven paradigms achieve competence in Mild ARDS scenarios—exemplified by Reflexion attaining an oxygenation reward ($r_{oxy}$) of 2825.64—their efficacy degrades precipitously as patient severity escalates. This trend highlights the limitation of monolithic reasoning architectures in navigating the adversarial trade-offs between oxygenation, acid-base homeostasis ($r_{vent}$), and lung mechanics ($r_{mech}$) under high physiological heterogeneity.

In the challenging \textbf{Severe} patient cohort, VentAgent establishes state-of-the-art performance, achieving a cumulative oxygenation reward of \textbf{2240.33}. This significantly surpasses the most robust ensemble baseline, Tree of Thoughts (ToT), which plateaus at 1955.15. 
The performance divergence is most critical in the ventilation metric ($r_{vent}$), which penalizes deviations in pH and PaCO$_2$. Here, standard prompting strategies exhibit catastrophic failure modes; Linear CoT and Few-Shot methods regress to deeply negative rewards of \textbf{-128.24} and \textbf{-142.56}, respectively. Even advanced feedback mechanisms like Reflexion fail to arrest this decline (-102.37), suggesting that iterative verbal reinforcement alone cannot compensate for the lack of structured clinical priors. VentAgent, conversely, demonstrates superior stability, maintaining a ventilation score of \textbf{-42.58}, thereby minimizing iatrogenic risk while sustaining gas exchange.

This resilience extends to mechanical safety ($r_{mech}$), where VentAgent secures a reward of \textbf{175.64}, substantially exceeding standard CoT (95.67). The systemic superiority over methods like ReAct and Debate is attributable to the \textit{hierarchical decoupling} of the decision process. Unlike "black-box" optimization that attempts to map high-dimensional states directly to actions, VentAgent leverages heterogeneous experts to explicitly delineate the Pareto frontier, successfully navigating the non-linear physiological constraints of severe ARDS.

\begin{table*}[!t]
\centering
\renewcommand{\arraystretch}{1.2}
\setlength{\tabcolsep}{1.5pt}
\caption{Ablation study visualization. The layout is simplified by removing vertical lines and reducing horizontal separators. Whitespace is used to group related experiments.}
\label{tab:ablation_study}
\resizebox{\textwidth}{!}{
\begin{tabular}{lccccccccc}
\toprule
\multirow{2.5}{*}{\textbf{Method}} & \multicolumn{3}{c}{\textbf{Mild}} & \multicolumn{3}{c}{\textbf{Moderate}} & \multicolumn{3}{c}{\textbf{Severe}} \\
\cmidrule(lr){2-4} \cmidrule(lr){5-7} \cmidrule(lr){8-10}
 & \textbf{$r_{oxy}$} & \textbf{$r_{vent}$} & \textbf{$r_{mech}$} & \textbf{$r_{oxy}$} & \textbf{$r_{vent}$} & \textbf{$r_{mech}$} & \textbf{$r_{oxy}$} & \textbf{$r_{vent}$} & \textbf{$r_{mech}$} \\
\midrule
\textbf{Full VentAgent (Ours)} 
 & \grad{25}{\textbf{2865.60}}{} 
 & \grad{25}{\textbf{332.55}}{} 
 & \grad{22}{\textbf{1016.35}}{} 
 & \grad{22}{\textbf{2630.60}}{} 
 & \grad{20}{\textbf{153.80}}{} 
 & \grad{22}{\textbf{563.60}}{} 
 & \grad{25}{\textbf{2181.45}}{} 
 & \grad{22}{\textbf{-36.77}}{} 
 & \grad{20}{\textbf{161.20}}{} \\
w/o Perception Layer 
 & \grad{15}{2728.85}{\fall{-136.75}} 
 & \grad{20}{312.95}{\fall{-19.60}} 
 & \grad{15}{931.65}{\fall{-84.70}} 
 & \grad{12}{2496.30}{\fall{-134.30}} 
 & \grad{15}{115.70}{\fall{-38.10}} 
 & \grad{12}{427.10}{\fall{-136.50}} 
 & \grad{10}{1927.85}{\fall{-253.60}} 
 & \grad{15}{-70.87}{\fall{-34.10}} 
 & \grad{8}{77.20}{\fall{-84.00}} \\

\multicolumn{10}{l}{\textbf{\textit{w/o Planning Layer}}} \\
\quad w/o Scenario Meta-Agent 
 & \grad{22}{2816.80}{\fall{-48.80}} 
 & \grad{22}{308.45}{\fall{-24.10}} 
 & \grad{20}{984.75}{\fall{-31.60}} 
 & \grad{20}{2560.30}{\fall{-70.30}} 
 & \grad{15}{106.80}{\fall{-47.00}} 
 & \grad{20}{512.50}{\fall{-51.10}} 
 & \grad{20}{2066.35}{\fall{-115.10}} 
 & \grad{12}{-88.77}{\fall{-52.00}} 
 & \grad{18}{129.80}{\fall{-31.40}} \\

\quad w/o Domain Experts 
 & \grad{20}{2799.50}{\fall{-66.10}} 
 & \grad{20}{295.55}{\fall{-37.00}} 
 & \grad{20}{975.75}{\fall{-40.60}} 
 & \grad{18}{2524.80}{\fall{-105.80}} 
 & \grad{18}{119.80}{\fall{-34.00}} 
 & \grad{20}{516.60}{\fall{-47.00}} 
 & \grad{18}{2030.85}{\fall{-150.60}} 
 & \grad{15}{-79.57}{\fall{-42.80}} 
 & \grad{18}{138.30}{\fall{-22.90}} \\

\multicolumn{10}{l}{\textbf{\textit{w/o Orchestration Layer}}} \\
\quad w/o Pulm. Protection 
 & \grad{25}{2856.50}{\fall{-9.10}} 
 & \grad{24}{326.05}{\fall{-6.50}} 
 & \grad{12}{923.15}{\fall{-93.20}} 
 & \grad{42}{2690.58}{\rise{60.98}} 
 & \grad{18}{141.90}{\fall{-11.90}} 
 & \grad{8}{340.80}{\fall{-222.80}} 
 & \grad{45}{2295.55}{\rise{114.10}} 
 & \grad{20}{-39.77}{\fall{-3.00}} 
 & \grad{5}{68.30}{\fall{-92.90}} \\

\quad w/o Oxy-Security 
 & \grad{8}{2649.80}{\fall{-215.80}} 
 & \grad{42}{340.55}{\rise{8.00}} 
 & \grad{45}{1050.25}{\rise{33.90}} 
 & \grad{8}{2348.90}{\fall{-281.70}} 
 & \grad{35}{170.52}{\rise{16.72}} 
 & \grad{40}{610.55}{\rise{46.95}} 
 & \grad{5}{1716.45}{\fall{-465.00}} 
 & \grad{38}{-25.55}{\rise{11.22}} 
 & \grad{42}{210.42}{\rise{49.22}} \\

\quad w/o Acid-Base Homeost. 
 & \grad{28}{2870.55}{\rise{4.95}} 
 & \grad{10}{172.15}{\fall{-160.40}} 
 & \grad{22}{993.55}{\fall{-22.80}} 
 & \grad{28}{2640.25}{\rise{9.65}} 
 & \grad{5}{-21.60}{\fall{-175.40}} 
 & \grad{22}{543.00}{\fall{-20.60}} 
 & \grad{24}{2154.05}{\fall{-27.40}} 
 & \grad{5}{-163.27}{\fall{-126.50}} 
 & \grad{24}{170.25}{\rise{9.05}} \\

w/o Sparse Reflective Memory 
 & \grad{18}{2764.10}{\fall{-101.50}} 
 & \grad{18}{295.65}{\fall{-36.90}} 
 & \grad{20}{993.35}{\fall{-23.00}} 
 & \grad{18}{2533.60}{\fall{-97.00}} 
 & \grad{18}{129.00}{\fall{-24.80}} 
 & \grad{20}{525.70}{\fall{-37.90}} 
 & \grad{18}{2092.55}{\fall{-88.90}} 
 & \grad{18}{-53.17}{\fall{-16.40}} 
 & \grad{18}{143.30}{\fall{-17.90}} \\

w/o Hierarchical Audit
 & \grad{20}{2795.40}{\fall{-70.20}}
 & \grad{20}{303.95}{\fall{-28.60}}
 & \grad{20}{998.35}{\fall{-18.00}}
 & \grad{20}{2564.30}{\fall{-66.30}}
 & \grad{20}{135.00}{\fall{-18.80}}
 & \grad{20}{539.10}{\fall{-24.50}}
 & \grad{20}{2114.75}{\fall{-66.70}}
 & \grad{20}{-49.17}{\fall{-12.40}}
 & \grad{20}{149.20}{\fall{-12.00}} \\

\bottomrule
\end{tabular}
}
\end{table*}

\subsection{Ablation Study}
\label{sec:ablation}

To validate the necessity of each component in the VentAgent hierarchical framework, we conducted a comprehensive ablation study (Table \ref{tab:ablation_study}). The results demonstrate that performance stems from the synergistic integration of semantic awareness, decoupled planning, and multi-perspective coordination.

\noindent\textbf{Impact of Semantic Perception on Safety Boundaries.} 
Removing the \textit{Perception Layer} results in a systemic collapse, with the most profound deterioration observed in the Severe cohort ($r_{oxy}$ decreases by 253.60 and $r_{mech}$ by 84.00). This confirms that raw high-dimensional telemetry alone is insufficient. Without the explicit state diagnosis provided by $G_{percept}$, the agent loses its "semantic grounding," degenerating into a reactive controller unable to identify the safe operational boundaries ($\mathcal{R}_{safe}$).

\noindent\textbf{Necessity of Strategic Decoupling in Planning.} 
Ablating the \textit{Scenario Meta-Agent} leads to a significant drop in ventilation rewards ($r_{vent}$ decreases by 52.00 in Severe cases). This validates our hypothesis that a strategic intent (e.g., "rescue" phase) must precede specific parameter adjustments. Similarly, removing \textit{Domain Experts} constrains the exploration space, causing a consistent decline in oxygenation scores ($r_{oxy}$ drops by 150.60), proving that a monolithic planner cannot effectively navigate the combinatorial complexity of the action space.

\noindent\textbf{Orchestration as a Safety Valve.} 
The ablation of the \textit{Orchestration Layer} provides strong evidence for our arbitration mechanism. 
(1) \textit{Lung Protection:} Removing the Pulmonary Protection Assessor causes mechanical safety to plummet ($r_{mech}$ drops by 222.80 in Moderate cases), as the system aggressively over-ventilates to maximize blood gas metrics. 
(2) \textit{Oxygenation Security:} Conversely, removing the Oxy-Security Assessor leads to a massive failure in oxygenation ($r_{oxy}$ drops by 465.00 in Severe cases), highlighting that standard protocols cannot stabilize life-threatening hypoxemia without a dedicated safety override.

\noindent\textbf{Long-term Consistency and Medical Integrity.} 
The exclusion of \textit{Sparse Reflective Memory} leads to consistent performance degradation (Severe $r_{oxy}$ drops by 88.90), underscoring the need for temporal context to resolve sub-optimal trends like CO$_2$ retention. Furthermore, removing the \textit{Hierarchical Audit} module results in pervasive attenuation across all metrics (Severe $r_{oxy}$ declines by 66.70), highlighting the critical role of run-time verification in preventing sporadic reasoning hallucinations.

\begin{figure}[t]
  \centering
  \includegraphics[width=0.85\linewidth]{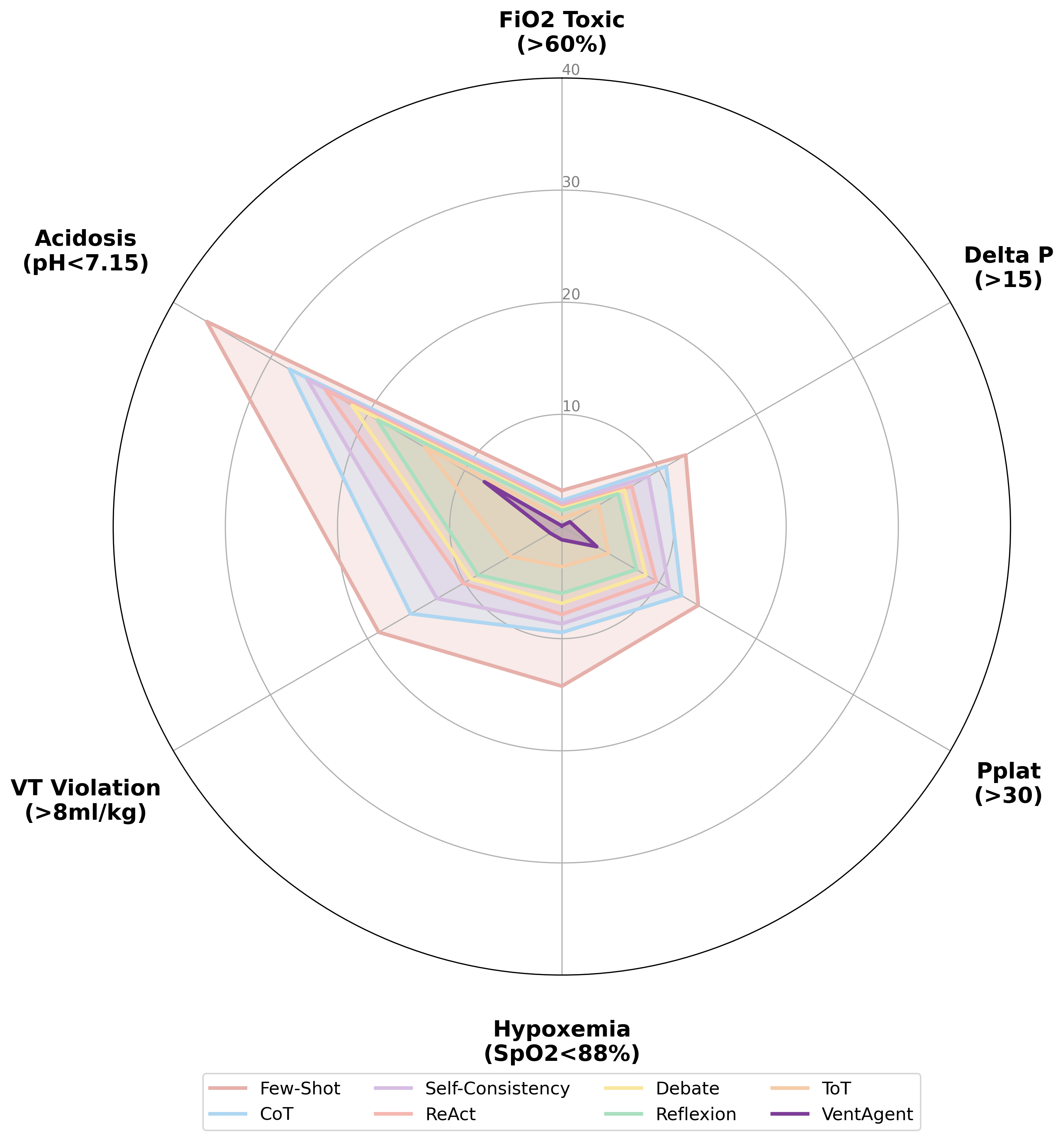} 
  \caption{ Comparative analysis of clinical safety violations. VentAgent (blue) minimizes critical violations compared to CoT (orange) and Few-Shot (green) baselines.}
  \label{fig:safety_violations}
\end{figure}

\subsection{Clinical Safety and Constraint Analysis} 

We rigorously assess the clinical reliability of the proposed framework by analyzing constraint compliance (Figure \ref{fig:safety_violations}). \textbf{VentAgent} exhibits a systematic advantage in maintaining physiological safety boundaries. In the high-risk domain of severe ARDS, baseline methods such as Few-Shot and CoT manifest high failure rates, exhibiting respiratory acidosis violation rates ($pH < 7.15$) as high as 36.57\% and 28.06\%, respectively. In contrast, VentAgent effectively delineates the Pareto frontier, reducing this risk to 7.98\%. Simultaneously, it suppresses ventilator-induced lung injury (VILI) risks \citep{slutsky2013ventilator}, maintaining Tidal Volume ($V_T$) violations at a negligible 1.19\% compared to the 18.85\% failure rate in Few-Shot approaches.

Crucially, the Arbitration mechanism prevents the conflation of "hard" limits and "soft" targets. Standard baselines frequently treat a violation of Driving Pressure ($\Delta P > 15$) as a tradable penalty. The high $\Delta P$ violation rate in baselines (12.75\%) confirms this alignment failure. Conversely, VentAgent’s hierarchical veto mechanism enforces mechanical safety unconditionally.

\subsection{Mechanics of Self-Correction via Dynamic Intervention} 

\begin{table}[t]
\centering
\caption{Quantitative Analysis of Audit Interventions. We report the Correction Trigger Rate (CTR), the distribution of intercepted error types, and the Safety Recovery Rate (SRR). Notably, error distribution shifts from \textit{Completeness} issues in mild cases to \textit{Logical} inconsistencies in severe scenarios.}
\label{tab:audit_analysis}
\resizebox{\columnwidth}{!}{%
\begin{tabular}{llccccc}
\toprule
\multirow{2}{*}{\textbf{Severity}} & \multirow{2}{*}{\textbf{Stage}} & \multirow{2}{*}{\textbf{CTR (\%)}} & \multicolumn{3}{c}{\textbf{Error Type Distribution (\%)}} & \multirow{2}{*}{\textbf{SRR (\%)}} \\
\cmidrule(lr){4-6}
 &  &  & \textbf{Fact.} & \textbf{Logic.} & \textbf{Comp.} &  \\
\midrule
\multirow{3}{*}{\textbf{Mild}} & Percept & 1.2 & 85.4 & 8.2 & 6.4 & 98.5 \\
 & Plan & 2.5 & 12.1 & 15.3 & 72.6 & 99.1 \\
 & Orch & 1.1 & 5.5 & 12.4 & 82.1 & 99.0 \\
\midrule
\multirow{3}{*}{\textbf{Mod}} & Percept & 3.4 & 76.2 & 14.5 & 9.3 & 96.2 \\
 & Plan & 6.8 & 15.6 & 62.4 & 22.0 & 95.8 \\
 & Orch & 2.9 & 10.2 & 58.7 & 31.1 & 97.4 \\
\midrule
\multirow{3}{*}{\textbf{Severe}} & Percept & 4.5 & 68.9 & 21.5 & 9.6 & 92.1 \\
 & \textbf{Plan} & \textbf{11.5} & 11.2 & \textbf{78.4} & 10.4 & \textbf{94.2} \\
 & Orch & 4.2 & 8.5 & 35.1 & \textbf{56.4} & 95.5 \\
\bottomrule
\end{tabular}%
}
\end{table}

We conducted a fine-grained analysis to quantify how VentAgent mitigates LLM stochasticity (Table \ref{tab:audit_analysis}).
\textbf{1) Severity-Dependent Activation:} The Correction Trigger Rate (CTR) scales monotonically with patient severity. While the module remains quiescent in Mild cases (CTR $\approx$ 1.2\%), activation surges to 11.5\% during Severe Planning. This reflects the escalating difficulty of adversarial trade-offs in critical states.
\textbf{2) Shift in Error Typology:} There is a distinct distributional shift in error modes. In Mild scenarios, interventions primarily address \textit{Completeness} (72.6\%), suggesting procedural omissions. Conversely, under Severe conditions, the dominant error mode transitions to \textit{Logical Coherence} (peaking at 78.4\%). Here, the Audit Agent actively rectifies contradictions, such as proposed volume increases that violate a "stiff lung" diagnosis.
\textbf{3) Safety Assurance:} Despite the complexity, the Safety Recovery Rate (SRR) remains robust ($>92\%$ across all strata), guaranteeing that robustness stems from active run-time self-correction.

\subsection{Multi-Perspective Trade-off Analysis}

\begin{figure}[t]
  \centering
  \includegraphics[width=\linewidth]{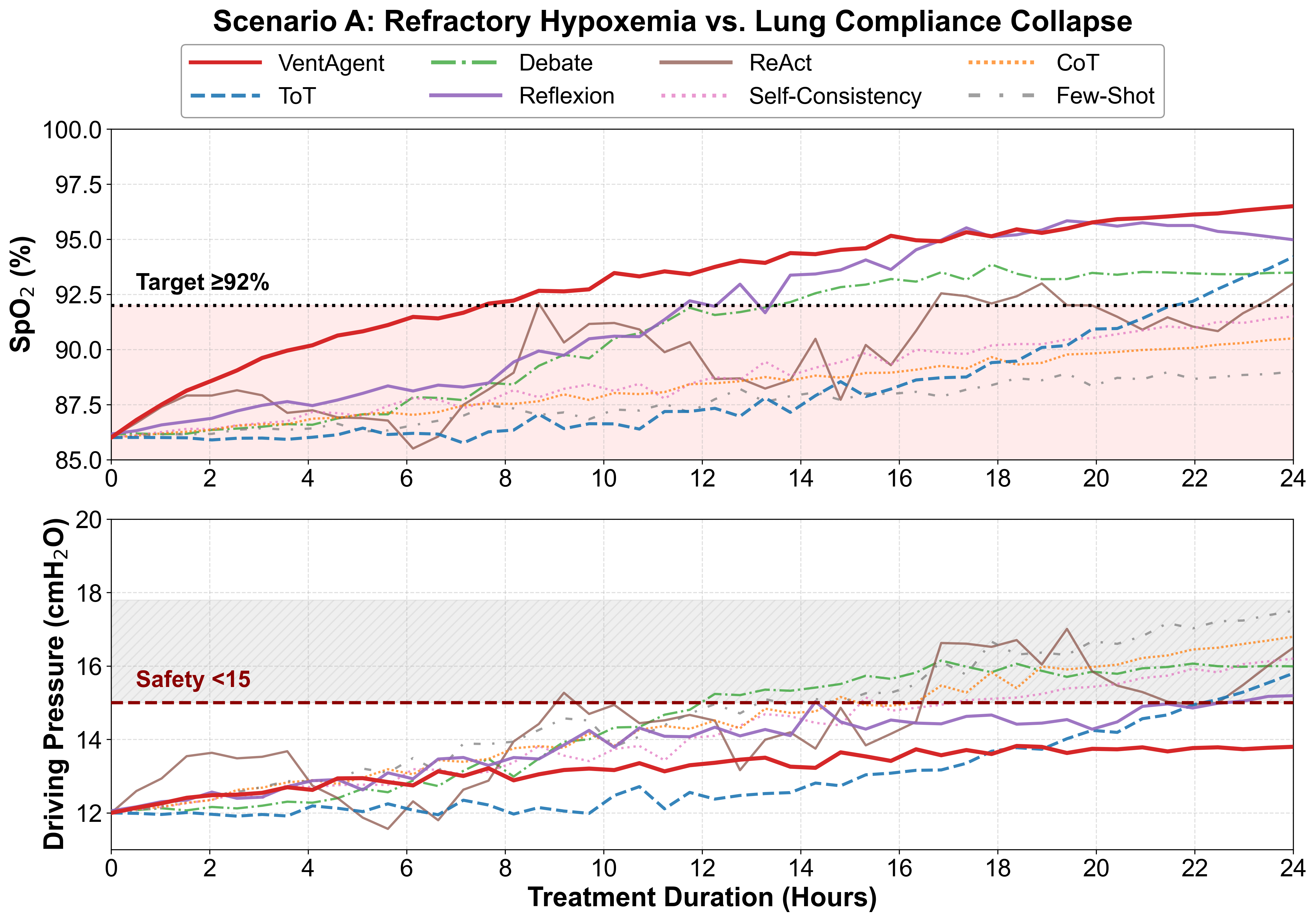} \\
  \vspace{0.5em} 
  \includegraphics[width=\linewidth]{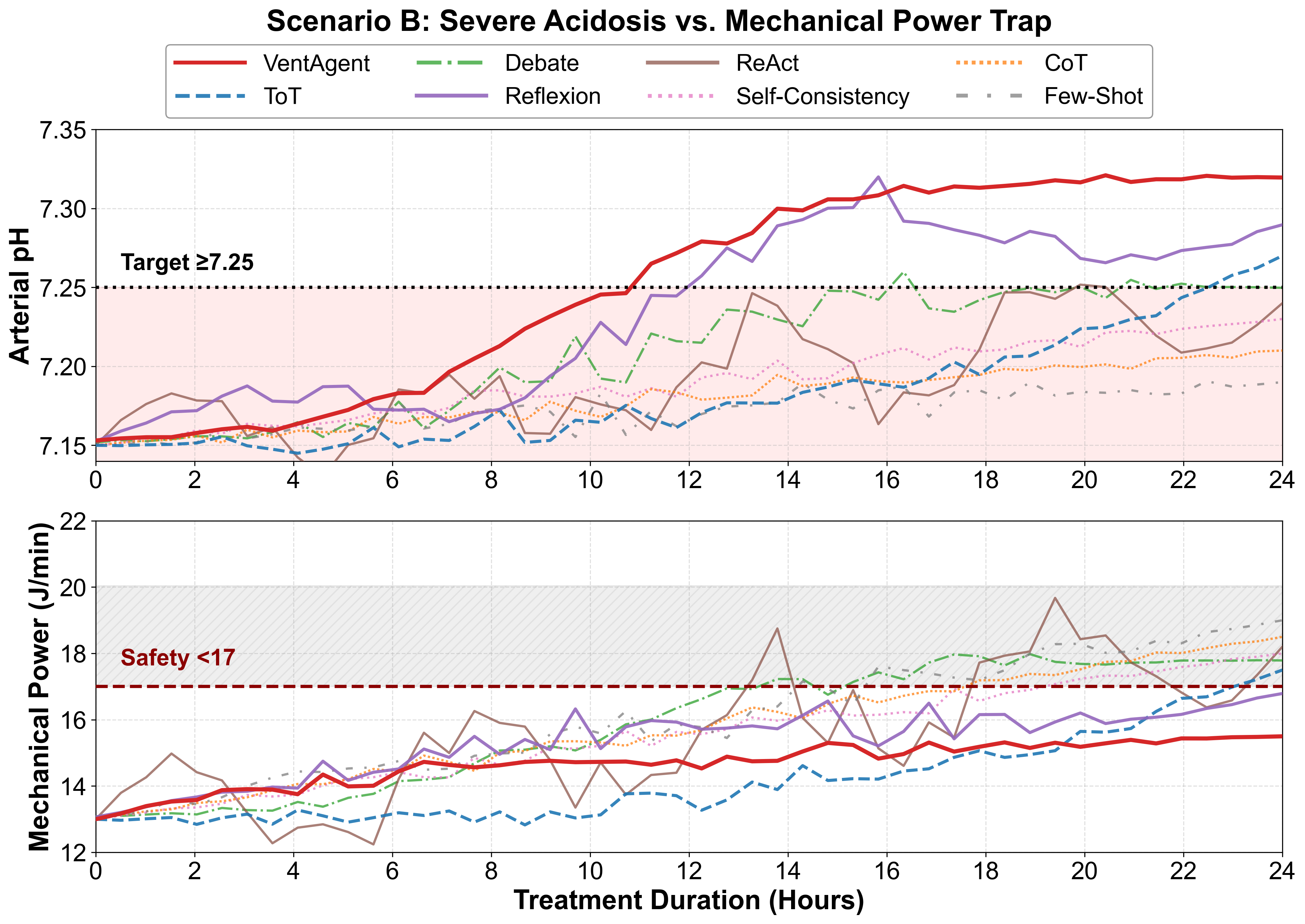} 
  \caption{ Resolution of clinical deadlocks. (Top) Scenario A: VentAgent maintains safe driving pressure (blue) unlike ToT (orange). (Bottom) Scenario B: VentAgent stabilizes mechanical power via permissive hypercapnia.}
  \label{fig:scenario}
\end{figure}

Aggregated metrics often obscure the nuance of "clinical deadlocks." We analyze two specific scenarios to demonstrate VentAgent's capacity to navigate adversarial trade-offs (Figure~\ref{fig:scenario}).

\noindent\textbf{Scenario A: Resolution of Hypoxemia-Compliance Conflict.}
In severe ARDS, aggressive recruitment risks barotrauma. While baselines (e.g., ReAct, ToT) exhibit stochastic volatility and frequently breach the Driving Pressure safety threshold ($\Delta P > 15$ cmH$_2$O), VentAgent adheres strictly to a lung-protective strategy. It maintains $\Delta P$ consistently within a safe range ($12$--$14$ cmH$_2$O), yet successfully achieves the therapeutic oxygenation target ($SpO_2 \geq 92\%$) by Hour 9. This trajectory validates the Orchestration Layer's efficacy in prioritizing mechanical safety over high-risk recruitment.

\noindent\textbf{Scenario B: Strategic Management of the "Power Trap".}
Addressing severe respiratory acidosis ($pH < 7.15$) often tempts a compensatory increase in ventilation, escalating Mechanical Power ($P_{pow}$) and VILI risk \citep{gattinoni2016ventilator, slutsky2013ventilator}. VentAgent deploys a \textit{permissive hypercapnia} strategy, stabilizing $P_{pow}$ within a "Safety Corridor" ($\approx 15.5$ J/min) \citep{hickling1994low}. Unlike baselines that pursue rapid pH normalization at the cost of dangerous power surges ($> 18$ J/min), VentAgent accepts a more gradual pH recovery (crossing 7.25 at Hour 11), effectively preventing ventilator-induced lung injury.
\section{Conclusion and Future Work}

In this work, we presented \textsc{VentAgent}, a hierarchical multi-agent framework that fundamentally reimagines physiological control not as a black-box optimization task, but as a transparent, multi-objective arbitration process. By operationalizing the principle of \textit{cognitive decoupling}, our framework successfully emulates the "Perception-Planning-Orchestration" clinical workflow, enabling Large Language Models to navigate the adversarial trade-offs between oxygenation, lung protection, and acid-base homeostasis with unprecedented interpretability.
Extensive evaluations on high-fidelity simulators demonstrate that \textsc{VentAgent} significantly outperforms state-of-the-art baselines, establishing a new paradigm for safe, consistent, and clinically aligned critical care automation. We believe this shift from opaque scalar rewards to structured semantic reasoning serves as a cornerstone for the next generation of trustworthy Medical AI.

Looking ahead, our research trajectory focuses on bridging the gap between algorithmic reasoning and physical deployment:
\begin{itemize}
    \item \textbf{Hardware-in-the-Loop (HIL) Validation:} To move beyond digital simulation, we plan to deploy \textsc{VentAgent} on a physical test lung platform (e.g., ASL 5000). This will validate the system's robustness against real-world sensor noise, actuator latency, and non-linear physical respiratory mechanics that pure software simulators may idealize.
    \item \textbf{Adaptive Agent Collaboration:} We aim to refine the structural efficiency of the multi-agent framework. Future iterations will explore dynamic coalition formation, where the agent topology (e.g., the number of active experts) adapts in real-time to patient severity. This optimization intends to reduce token latency and computational overhead without compromising decision quality.
    \item \textbf{Clinical Data Alignment:} Finally, we will extend our evaluation to retrospective large-scale datasets (e.g., MIMIC-IV) to further verify the generalization capability of our arbitration logic across diverse human demographics and pathological subtypes.
\end{itemize}

\section{Ethical Considerations}
\textsc{VentAgent} functions strictly as a \textit{human-in-the-loop} CDSS, not an autonomous substitute. Given inherent LLM stochasticity, clinical deployment mandates immutable hardware-level safety constraints (e.g., pressure valves) that unconditionally supersede algorithmic commands. Furthermore, continuous fairness auditing is imperative to detect latent biases in pre-trained models, ensuring equitable care across diverse patient demographics.

\bibliographystyle{ACM-Reference-Format}
\bibliography{references}

@article{ranieri2012acute,
  title={Acute respiratory distress syndrome: the Berlin Definition.},
  author={Ranieri, V Marco and Rubenfeld, Gordon D and Taylor Thompson, B and Ferguson, Niall D and Caldwell, Ellen and Fan, Eddy and Camporota, Luigi and Slutsky, Arthur S},
  journal={JAMA: Journal of the American Medical Association},
  volume={307},
  number={23},
  year={2012}
}

@article{acute2000ventilation,
  title={Ventilation with lower tidal volumes as compared with traditional tidal volumes for acute lung injury and the acute respiratory distress syndrome},
  author={Acute Respiratory Distress Syndrome Network},
  journal={New England Journal of Medicine},
  volume={342},
  number={18},
  pages={1301--1308},
  year={2000},
  publisher={Mass Medical Soc}
}

@article{calfee2014subphenotypes,
  title={Subphenotypes in acute respiratory distress syndrome: latent class analysis of data from two randomised controlled trials},
  author={Calfee, Carolyn S and Delucchi, Kevin and Parsons, Polly E and Thompson, B Taylor and Ware, Lorraine B and Matthay, Michael A},
  journal={The Lancet Respiratory Medicine},
  volume={2},
  number={8},
  pages={611--620},
  year={2014},
  publisher={Elsevier}
}

@article{gottesman2019guidelines,
  title={Guidelines for reinforcement learning in healthcare},
  author={Gottesman, Omer and Johansson, Fredrik and Komorowski, Matthieu and Faisal, Aldo and Sontag, David and Doshi-Velez, Finale and Celi, Leo Anthony},
  journal={Nature medicine},
  volume={25},
  number={1},
  pages={16--18},
  year={2019},
  publisher={Nature Publishing Group US New York}
}

@article{wornow2023shaky,
  title={The shaky foundations of large language models and foundation models for electronic health records},
  author={Wornow, Michael and Xu, Yizhe and Thapa, Rahul and Patel, Birju and Steinberg, Ethan and Fleming, Scott and Pfeffer, Michael A and Fries, Jason and Shah, Nigam H},
  journal={npj digital medicine},
  volume={6},
  number={1},
  pages={135},
  year={2023},
  publisher={Nature Publishing Group UK London}
}

@article{levine2020offline,
  title={Offline reinforcement learning: Tutorial, review, and perspectives on open problems},
  author={Levine, Sergey and Kumar, Aviral and Tucker, George and Fu, Justin},
  journal={arXiv preprint arXiv:2005.01643},
  year={2020}
}

@article{richens2020improving,
  title={Improving the accuracy of medical diagnosis with causal machine learning},
  author={Richens, Jonathan G and Lee, Ciar{\'a}n M and Johri, Saurabh},
  journal={Nature communications},
  volume={11},
  number={1},
  pages={3923},
  year={2020},
  publisher={Nature Publishing Group UK London}
}

@article{komorowski2018artificial,
  title={The artificial intelligence clinician learns optimal treatment strategies for sepsis in intensive care},
  author={Komorowski, Matthieu and Celi, Leo A and Badawi, Omar and Gordon, Anthony C and Faisal, A Aldo},
  journal={Nature medicine},
  volume={24},
  number={11},
  pages={1716--1720},
  year={2018},
  publisher={Nature Publishing Group US New York}
}

@article{bray2019pulse,
  title={Pulse physiology engine: an open-source software platform for computational modeling of human medical simulation},
  author={Bray, Aaron and Webb, Jeffrey B and Enquobahrie, Andinet and Vicory, Jared and Heneghan, Jerry and Hubal, Robert and TerMaath, Stephanie and Asare, Philip and Clipp, Rachel B},
  journal={SN Comprehensive Clinical Medicine},
  volume={1},
  number={5},
  pages={362--377},
  year={2019},
  publisher={Springer}
}

@article{raghu2017deep,
  title={Deep reinforcement learning for sepsis treatment},
  author={Raghu, Aniruddh and Komorowski, Matthieu and Ahmed, Imran and Celi, Leo and Szolovits, Peter and Ghassemi, Marzyeh},
  journal={arXiv preprint arXiv:1711.09602},
  year={2017}
}

@inproceedings{ward2024optimal,
  title={Optimal Control of Mechanical Ventilators with Learned Respiratory Dynamics},
  author={Ward, Isaac R and Asmar, Dylan M and Arief, Mansur and Mike, Jana Krystofova and Kochenderfer, Mykel J},
  booktitle={2024 IEEE 37th International Symposium on Computer-Based Medical Systems (CBMS)},
  pages={192--198},
  year={2024},
  organization={IEEE}
}

@article{singhal2023large,
  title={Large language models encode clinical knowledge},
  author={Singhal, Karan and Azizi, Shekoofeh and Tu, Tao and Mahdavi, S Sara and Wei, Jason and Chung, Hyung Won and Scales, Nathan and Tanwani, Ajay and Cole-Lewis, Heather and Pfohl, Stephen and others},
  journal={Nature},
  volume={620},
  number={7972},
  pages={172--180},
  year={2023},
  publisher={Nature Publishing Group}
}

@article{nori2023capabilities,
  title={Capabilities of gpt-4 on medical challenge problems},
  author={Nori, Harsha and King, Nicholas and McKinney, Scott Mayer and Carignan, Dean and Horvitz, Eric},
  journal={arXiv preprint arXiv:2303.13375},
  year={2023}
}

@article{wei2022chain,
  title={Chain-of-thought prompting elicits reasoning in large language models},
  author={Wei, Jason and Wang, Xuezhi and Schuurmans, Dale and Bosma, Maarten and Xia, Fei and Chi, Ed and Le, Quoc V and Zhou, Denny and others},
  journal={Advances in neural information processing systems},
  volume={35},
  pages={24824--24837},
  year={2022}
}

@inproceedings{yao2022react,
  title={React: Synergizing reasoning and acting in language models},
  author={Yao, Shunyu and Zhao, Jeffrey and Yu, Dian and Du, Nan and Shafran, Izhak and Narasimhan, Karthik R and Cao, Yuan},
  booktitle={The eleventh international conference on learning representations},
  year={2022}
}

@article{xi2025rise,
  title={The rise and potential of large language model based agents: A survey},
  author={Xi, Zhiheng and Chen, Wenxiang and Guo, Xin and He, Wei and Ding, Yiwen and Hong, Boyang and Zhang, Ming and Wang, Junzhe and Jin, Senjie and Zhou, Enyu and others},
  journal={Science China Information Sciences},
  volume={68},
  number={2},
  pages={121101},
  year={2025},
  publisher={Springer}
}

@article{tu2024towards,
  title={Towards generalist biomedical AI},
  author={Tu, Tao and Azizi, Shekoofeh and Driess, Danny and Schaekermann, Mike and Amin, Mohamed and Chang, Pi-Chuan and Carroll, Andrew and Lau, Charles and Tanno, Ryutaro and Ktena, Ira and others},
  journal={Nejm Ai},
  volume={1},
  number={3},
  pages={AIoa2300138},
  year={2024},
  publisher={Massachusetts Medical Society}
}

@article{johnson2016mimic,
  title={MIMIC-III, a freely accessible critical care database},
  author={Johnson, Alistair EW and Pollard, Tom J and Shen, Lu and Lehman, Li-wei H and Feng, Mengling and Ghassemi, Mohammad and Moody, Benjamin and Szolovits, Peter and Anthony Celi, Leo and Mark, Roger G},
  journal={Scientific data},
  volume={3},
  number={1},
  pages={1--9},
  year={2016},
  publisher={Nature Publishing Group}
}

@article{johnson2023mimic,
  title={MIMIC-IV, a freely accessible electronic health record dataset},
  author={Johnson, Alistair EW and Bulgarelli, Lucas and Shen, Lu and Gayles, Alvin and Shammout, Ayad and Horng, Steven and Pollard, Tom J and Hao, Sicheng and Moody, Benjamin and Gow, Brian and others},
  journal={Scientific data},
  volume={10},
  number={1},
  pages={1},
  year={2023},
  publisher={Nature Publishing Group UK London}
}

@inproceedings{kondrup2023towards,
  title={Towards safe mechanical ventilation treatment using deep offline reinforcement learning},
  author={Kondrup, Flemming and Jiralerspong, Thomas and Lau, Elaine and de Lara, Nathan and Shkrob, Jacob and Tran, My Duc and Precup, Doina and Basu, Sumana},
  booktitle={Proceedings of the AAAI Conference on Artificial Intelligence},
  volume={37},
  number={13},
  pages={15696--15702},
  year={2023}
}

@article{prasad2017reinforcement,
  title={A reinforcement learning approach to weaning of mechanical ventilation in intensive care units},
  author={Prasad, Niranjani and Cheng, Li-Fang and Chivers, Corey and Draugelis, Michael and Engelhardt, Barbara E},
  journal={arXiv preprint arXiv:1704.06300},
  year={2017}
}

@article{yu2021reinforcement,
  title={Reinforcement learning in healthcare: A survey},
  author={Yu, Chao and Liu, Jiming and Nemati, Shamim and Yin, Guosheng},
  journal={ACM Computing Surveys (CSUR)},
  volume={55},
  number={1},
  pages={1--36},
  year={2021},
  publisher={ACM New York, NY}
}

@article{futoma2020popcorn,
  title={Popcorn: Partially observed prediction constrained reinforcement learning},
  author={Futoma, Joseph and Hughes, Michael C and Doshi-Velez, Finale},
  journal={arXiv preprint arXiv:2001.04032},
  year={2020}
}

@inproceedings{oberst2019counterfactual,
  title={Counterfactual off-policy evaluation with gumbel-max structural causal models},
  author={Oberst, Michael and Sontag, David},
  booktitle={International Conference on Machine Learning},
  pages={4881--4890},
  year={2019},
  organization={PMLR}
}

@inproceedings{jiang2016doubly,
  title={Doubly robust off-policy value evaluation for reinforcement learning},
  author={Jiang, Nan and Li, Lihong},
  booktitle={International conference on machine learning},
  pages={652--661},
  year={2016},
  organization={PMLR}
}

@inproceedings{thomas2016data,
  title={Data-efficient off-policy policy evaluation for reinforcement learning},
  author={Thomas, Philip and Brunskill, Emma},
  booktitle={International conference on machine learning},
  pages={2139--2148},
  year={2016},
  organization={PMLR}
}

@article{roijers2013survey,
  title={A survey of multi-objective sequential decision-making},
  author={Roijers, Diederik M and Vamplew, Peter and Whiteson, Shimon and Dazeley, Richard},
  journal={Journal of Artificial Intelligence Research},
  volume={48},
  pages={67--113},
  year={2013}
}

@article{garcia2015comprehensive,
  title={A comprehensive survey on safe reinforcement learning},
  author={Garc{\i}a, Javier and Fern{\'a}ndez, Fernando},
  journal={Journal of Machine Learning Research},
  volume={16},
  number={1},
  pages={1437--1480},
  year={2015}
}

@book{altman2021constrained,
  title={Constrained Markov decision processes},
  author={Altman, Eitan},
  year={2021},
  publisher={Routledge}
}

@article{amato2015driving,
  title={Driving pressure and survival in the acute respiratory distress syndrome},
  author={Amato, Marcelo BP and Meade, Maureen O and Slutsky, Arthur S and Brochard, Laurent and Costa, Eduardo LV and Schoenfeld, David A and Stewart, Thomas E and Briel, Matthias and Talmor, Daniel and Mercat, Alain and others},
  journal={New England Journal of Medicine},
  volume={372},
  number={8},
  pages={747--755},
  year={2015},
  publisher={Mass Medical Soc}
}

@inproceedings{achiam2017constrained,
  title={Constrained policy optimization},
  author={Achiam, Joshua and Held, David and Tamar, Aviv and Abbeel, Pieter},
  booktitle={International conference on machine learning},
  pages={22--31},
  year={2017},
  organization={PMLR}
}

@article{singhal2025toward,
  title={Toward expert-level medical question answering with large language models},
  author={Singhal, Karan and Tu, Tao and Gottweis, Juraj and Sayres, Rory and Wulczyn, Ellery and Amin, Mohamed and Hou, Le and Clark, Kevin and Pfohl, Stephen R and Cole-Lewis, Heather and others},
  journal={Nature Medicine},
  volume={31},
  number={3},
  pages={943--950},
  year={2025},
  publisher={Nature Publishing Group US New York}
}

@article{shen2023hugginggpt,
  title={Hugginggpt: Solving ai tasks with chatgpt and its friends in hugging face},
  author={Shen, Yongliang and Song, Kaitao and Tan, Xu and Li, Dongsheng and Lu, Weiming and Zhuang, Yueting},
  journal={Advances in Neural Information Processing Systems},
  volume={36},
  pages={38154--38180},
  year={2023}
}

@inproceedings{tang2024medagents,
  title={Medagents: Large language models as collaborators for zero-shot medical reasoning},
  author={Tang, Xiangru and Zou, Anni and Zhang, Zhuosheng and Li, Ziming and Zhao, Yilun and Zhang, Xingyao and Cohan, Arman and Gerstein, Mark},
  booktitle={Findings of the Association for Computational Linguistics: ACL 2024},
  pages={599--621},
  year={2024}
}

@article{ji2023survey,
  title={Survey of hallucination in natural language generation},
  author={Ji, Ziwei and Lee, Nayeon and Frieske, Rita and Yu, Tiezheng and Su, Dan and Xu, Yan and Ishii, Etsuko and Bang, Ye Jin and Madotto, Andrea and Fung, Pascale},
  journal={ACM computing surveys},
  volume={55},
  number={12},
  pages={1--38},
  year={2023},
  publisher={ACM New York, NY}
}

@article{fan2017official,
  title={An official American Thoracic Society/European Society of Intensive Care Medicine/Society of Critical Care Medicine clinical practice guideline: mechanical ventilation in adult patients with acute respiratory distress syndrome},
  author={Fan, Eddy and Del Sorbo, Lorenzo and Goligher, Ewan C and Hodgson, Carol L and Munshi, Laveena and Walkey, Allan J and Adhikari, Neill KJ and Amato, Marcelo BP and Branson, Richard and Brower, Roy G and others},
  journal={American journal of respiratory and critical care medicine},
  volume={195},
  number={9},
  pages={1253--1263},
  year={2017},
  publisher={American Thoracic Society}
}

@article{slutsky2013ventilator,
  title={Ventilator-induced lung injury},
  author={Slutsky, Arthur S and Ranieri, V Marco},
  journal={New England Journal of Medicine},
  volume={369},
  number={22},
  pages={2126--2136},
  year={2013},
  publisher={Mass Medical Soc}
}

@article{gattinoni2016ventilator,
  title={Ventilator-related causes of lung injury: the mechanical power},
  author={Gattinoni, Luciano and Tonetti, Tommaso and Cressoni, Massimo and Cadringher, Paolo and Herrmann, Peter and Moerer, Onnen and Protti, Alessandro and Gotti, Miriam and Chiurazzi, Chiara and Carlesso, Eleonora and others},
  journal={Intensive care medicine},
  volume={42},
  number={10},
  pages={1567--1575},
  year={2016},
  publisher={Springer}
}

@article{hickling1994low,
  title={Low mortality rate in adult respiratory distress syndrome using low-volume, pressure-limited ventilation with permissive hypercapnia: a prospective study},
  author={Hickling, Keith G and Walsh, John and Henderson, Seton and Jackson, Rodger},
  journal={Critical care medicine},
  volume={22},
  number={10},
  pages={1530--1539},
  year={1994},
  publisher={LWW}
}

@article{yao2023tree,
  title={Tree of thoughts: Deliberate problem solving with large language models},
  author={Yao, Shunyu and Yu, Dian and Zhao, Jeffrey and Shafran, Izhak and Griffiths, Tom and Cao, Yuan and Narasimhan, Karthik},
  journal={Advances in neural information processing systems},
  volume={36},
  pages={11809--11822},
  year={2023}
}

@article{wang2022self,
  title={Self-consistency improves chain of thought reasoning in language models},
  author={Wang, Xuezhi and Wei, Jason and Schuurmans, Dale and Le, Quoc and Chi, Ed and Narang, Sharan and Chowdhery, Aakanksha and Zhou, Denny},
  journal={arXiv preprint arXiv:2203.11171},
  year={2022}
}

@article{shinn2023reflexion,
  title={Reflexion: Language agents with verbal reinforcement learning},
  author={Shinn, Noah and Cassano, Federico and Gopinath, Ashwin and Narasimhan, Karthik and Yao, Shunyu},
  journal={Advances in Neural Information Processing Systems},
  volume={36},
  pages={8634--8652},
  year={2023}
}

@inproceedings{du2023improving,
  title={Improving factuality and reasoning in language models through multiagent debate},
  author={Du, Yilun and Li, Shuang and Torralba, Antonio and Tenenbaum, Joshua B and Mordatch, Igor},
  booktitle={Forty-first International Conference on Machine Learning},
  year={2023}
}

@article{madaan2023self,
  title={Self-refine: Iterative refinement with self-feedback},
  author={Madaan, Aman and Tandon, Niket and Gupta, Prakhar and Hallinan, Skyler and Gao, Luyu and Wiegreffe, Sarah and Alon, Uri and Dziri, Nouha and Prabhumoye, Shrimai and Yang, Yiming and others},
  journal={Advances in Neural Information Processing Systems},
  volume={36},
  pages={46534--46594},
  year={2023}
}

@article{bai2022constitutional,
  title={Constitutional ai: Harmlessness from ai feedback},
  author={Bai, Yuntao and Kadavath, Saurav and Kundu, Sandipan and Askell, Amanda and Kernion, Jackson and Jones, Andy and Chen, Anna and Goldie, Anna and Mirhoseini, Azalia and McKinnon, Cameron and others},
  journal={arXiv preprint arXiv:2212.08073},
  year={2022}
}

@article{brown2020language,
  title={Language models are few-shot learners},
  author={Brown, Tom and Mann, Benjamin and Ryder, Nick and Subbiah, Melanie and Kaplan, Jared D and Dhariwal, Prafulla and Neelakantan, Arvind and Shyam, Pranav and Sastry, Girish and Askell, Amanda and others},
  journal={Advances in neural information processing systems},
  volume={33},
  pages={1877--1901},
  year={2020}
}

@article{menick2024gpt,
  title={GPT-4o mini: advancing cost-efficient intelligence},
  author={Menick, Jacob and Lu, Kevin and Zhao, Shengjia and Wallace, E and Ren, H and Hu, H and Stathas, N and Such, F Petroski},
  journal={Open AI: San Francisco, CA, USA},
  year={2024}
}

@inproceedings{ng1999policy,
  title={Policy invariance under reward transformations: Theory and application to reward shaping},
  author={Ng, Andrew Y and Harada, Daishi and Russell, Stuart},
  booktitle={Icml},
  volume={99},
  pages={278--287},
  year={1999},
  organization={Citeseer}
}

\newpage
\appendix

\section{Definition of state space variables}

This section defines in detail the state space and action space variables used in the model interaction. To simulate realistic patient physiological responses, all state parameters are sampled and calculated in real time using the Pulse Physiology Engine. As shown in Table \ref{tab:variables_en}, the observation space encompasses key physiological indicators such as oxygenation status, ventilation adequacy, respiratory mechanics, and hemodynamics, while the action space corresponds to the adjustable parameters of the ventilator in VC-CMV mode.

\begin{table*}[htbp]
    \centering
    \small
    \setlength{\tabcolsep}{2pt}
    \renewcommand{\arraystretch}{1.1}
    \caption{Observation and Action Space Definitions}
    \label{tab:variables_en}
    \begin{tabular}{@{}>{\raggedright\arraybackslash}p{0.20\textwidth} >{\raggedright\arraybackslash}p{0.2\textwidth} >{\raggedright\arraybackslash}p{0.4\textwidth}@{}}
        \toprule
        \textbf{Variable} & \textbf{Units} & \textbf{Description} \\
        \midrule
        $\mathrm{SpO}_2$ & \% & Peripheral oxygen saturation (assesses gas exchange efficiency). \\
        $\mathrm{PaO}_2$ & mmHg & Partial pressure of oxygen in arterial blood. \\
        $\mathrm{PaO}_2/\mathrm{FiO}_2$ (P/F ratio) & mmHg & PaO$_2$-to-FiO$_2$ ratio (oxygenation severity index; a.k.a. Horowitz index). \\
        $Q_s/Q_t$ & \% & Pulmonary shunt fraction. \\
        pH & -- & Arterial blood pH (acidity or alkalinity). \\
        $\mathrm{PaCO}_2$ & mmHg & Partial pressure of carbon dioxide in arterial blood. \\
        $\mathrm{EtCO}_2$ & mmHg & End-tidal carbon dioxide partial pressure. \\
        $\mathrm{HCO}_3^-$ & mmol/L & Bicarbonate ion concentration. \\
        $\dot V_E$ & L/min & Minute ventilation (total expired/inspired ventilation per minute). \\
        $RR_{\mathrm{obs}}$ & breaths/min & Observed respiratory rate. \\
        $V_{T,\mathrm{obs}}$ & mL & Observed tidal volume. \\
        $P_{\mathrm{plat}}$ & cmH$_2$O & Plateau pressure. \\
        $P_{\mathrm{peak}}$ & cmH$_2$O & Peak inspiratory pressure. \\
        $\Delta P$ (driving pressure) & cmH$_2$O & Driving pressure ($\Delta P=P_{\mathrm{plat}}-\mathrm{PEEP}$). \\
        $P_{\mathrm{mean}}$ & cmH$_2$O & Mean airway pressure. \\
        $\mathrm{PEEP}_{\mathrm{tot}}$ & cmH$_2$O & Total PEEP (extrinsic + intrinsic/auto-PEEP). \\
        $C_{\mathrm{stat}}$ & mL/cmH$_2$O & Static respiratory system compliance. \\
        RSBI & breaths/min/L & Rapid shallow breathing index. \\
        I:E ratio & -- & Ratio of inspiratory to expiratory time. \\
        $R_{\mathrm{insp}}$ & cmH$_2$O\,s/L & Inspiratory resistance. \\
        $R_{\mathrm{exp}}$ & cmH$_2$O\,s/L & Expiratory resistance. \\
        HR & beats/min & Heart rate. \\
        MAP & mmHg & Mean arterial pressure. \\
        CO & L/min & Cardiac output. \\
        \midrule
        $\mathrm{PEEP}$ & cmH$_2$O & Positive end-expiratory pressure (regulates alveolar recruitment). \\
        $\mathrm{FiO}_2$ & \% & Fraction of inspired oxygen (often expressed as \% in ventilator settings). \\
        $RR_{\mathrm{set}}$ & breaths/min & Set respiratory rate. \\
        Set $V_T$ & mL/kg & Set tidal volume (normalized by ideal body weight). \\
        $T_{\mathrm{insp}}$ & s & Inspiratory time. \\
        Inspiratory flow & L/min & Set inspiratory flow rate. \\
        \bottomrule
    \end{tabular}
\end{table*}

\paragraph{State sampling strategy.}
The environment employs a passive, phase-aware sampling scheme aligned with the ventilator’s intrinsic breath-state transitions. During a post-action stabilization interval, the simulator advances without sampling to approach a new steady regime. Subsequently, dense sampling is performed at fixed temporal resolution $\Delta t$, while key mechanical variables are captured at physiologically meaningful transition points. Specifically, plateau pressure is sampled at the transition \textit{Pause} $\rightarrow$ \textit{Exhale} (when the engine computes pause-respiratory parameters), and total PEEP is sampled at \textit{Exhale} $\rightarrow$ \textit{Inhale} (end-expiratory pressure update). Continuous variables are accumulated over the sampling window and aggregated using a robust mean to mitigate outliers.

Let $\{x_i\}_{i=1}^n$ denote samples of a scalar variable $x$. A robust mean $\bar{x}_{\mathrm{rob}}$ is computed by median-based outlier rejection:
\[
m = \mathrm{median}(x_i), \quad
\mathrm{MAD} = \mathrm{median}\left(|x_i - m|\right),
\]
\[
z_i = 0.6745\,\frac{x_i - m}{\mathrm{MAD}}, \quad
\bar{x}_{\mathrm{rob}} = \frac{1}{|\mathcal{I}|}\sum_{i\in\mathcal{I}} x_i, \quad
\mathcal{I}=\{i:\ |z_i|\le 2.5\}.
\]
If fewer than three valid samples exist, the arithmetic mean is used; if all values are invalid, the estimate is reported as NaN.

The derived mechanical indices are computed from aggregated quantities. Driving pressure is
\[
\Delta P_{\mathrm{obs}} = \max\{0,\, P_{\mathrm{plat}} - \mathrm{PEEP}_{\mathrm{tot}}\},
\]
while an alveolar driving pressure is estimated by substituting the set PEEP:
\[
\Delta P_{\mathrm{alv}} = \max\{0,\, P_{\mathrm{plat}} - \mathrm{PEEP}_{\mathrm{set}}\}.
\]
When sufficient plateau samples exist, static compliance is recomputed as
\[
C_{\mathrm{rs}} = \frac{V_T}{P_{\mathrm{plat}} - \mathrm{PEEP}_{\mathrm{tot}}}
\quad\text{(mL/cmH$_2$O)},
\]
otherwise a fallback engine estimate is used. The aggregated values form the state vector, ensuring phase-consistent mechanical measurements and robust physiological summaries.

\section{Model-Based Evaluation and Data Curation Analysis}
\label{sec:appendix_rl_control}

\subsection{Computational Constraints and Dynamics Modeling}
Direct interaction with the high-fidelity Pulse Physiology Engine is computationally prohibitive for data-intensive optimization algorithms, requiring approximately 156 seconds to simulate a single integration step. To enable scalable training for Reinforcement Learning (RL) and control baselines, we employ the \textbf{Embed to Control (E2C)} architecture to learn a surrogate dynamics model. E2C utilizes a variational autoencoder to map high-dimensional physiological states into a compact latent embedding, coupled with a locally linear transition model that predicts state evolution. This surrogate environment reduces inference time to $\approx 2$ seconds per step, facilitating the high-throughput sampling required for iterative policy optimization.

\begin{table*}[t]
\centering
\small
\renewcommand{\arraystretch}{1.2} 
\setlength{\tabcolsep}{2.5pt}     
\caption{Performance comparison of RL and Control baselines.}
\label{tab:rl_control_results}
\begin{tabular}{ll ccc ccc ccc}
\toprule
\multirow{2.5}{*}{\textbf{Framework}} & \multirow{2.5}{*}{\textbf{Policy}} & \multicolumn{3}{c}{\textbf{Mild}} & \multicolumn{3}{c}{\textbf{Moderate}} & \multicolumn{3}{c}{\textbf{Severe}} \\
\cmidrule(lr){3-5} \cmidrule(lr){6-8} \cmidrule(lr){9-11}
 & & $r_{\text{oxy}}$ & $r_{\text{vent}}$ & $r_{\text{mech}}$ & $r_{\text{oxy}}$ & $r_{\text{vent}}$ & $r_{\text{mech}}$ & $r_{\text{oxy}}$ & $r_{\text{vent}}$ & $r_{\text{mech}}$ \\
\midrule

\multirow{3}{*}{\textbf{E2C-SMPC}} 
 & Random & \grad{5}{234.85}{} & \grad{5}{-77.43}{} & \grad{5}{24.45}{} & \grad{5}{129.20}{} & \grad{5}{-131.20}{} & \grad{5}{7.85}{} & \grad{5}{42.85}{} & \grad{5}{-470.44}{} & \grad{5}{-21.80}{} \\
 & Ardsnet & \grad{10}{1269.15}{} & \grad{5}{-25.27}{} & \grad{5}{91.85}{} & \grad{10}{990.25}{} & \grad{5}{-78.38}{} & \grad{5}{45.05}{} & \grad{10}{667.45}{} & \grad{5}{-366.46}{} & \grad{5}{7.50}{} \\
 & VentAgent & \grad{15}{1447.10}{} & \grad{10}{81.70}{} & \grad{10}{338.90}{} & \grad{15}{1162.25}{} & \grad{10}{21.70}{} & \grad{10}{178.45}{} & \grad{15}{815.45}{} & \grad{5}{-208.48}{} & \grad{5}{37.10}{} \\
\cmidrule(l){1-11} 

\multirow{3}{*}{\textbf{E2C-MPPI}} 
 & Random & \grad{5}{238.15}{} & \grad{5}{-75.72}{} & \grad{5}{24.40}{} & \grad{5}{141.45}{} & \grad{5}{-126.87}{} & \grad{5}{9.70}{} & \grad{5}{44.85}{} & \grad{5}{-457.57}{} & \grad{5}{-22.42}{} \\
 & Ardsnet & \grad{10}{1309.45}{} & \grad{5}{-24.70}{} & \grad{5}{94.50}{} & \grad{10}{1017.40}{} & \grad{5}{-76.76}{} & \grad{5}{45.40}{} & \grad{10}{691.15}{} & \grad{5}{-356.39}{} & \grad{5}{8.05}{} \\
 & VentAgent & \grad{15}{1487.05}{} & \grad{10}{83.60}{} & \grad{10}{348.55}{} & \grad{15}{1208.70}{} & \grad{10}{23.20}{} & \grad{10}{185.95}{} & \grad{15}{847.00}{} & \grad{5}{-202.35}{} & \grad{5}{37.70}{} \\
\cmidrule(l){1-11}

\multirow{3}{*}{\textbf{E2C-PPO}} 
 & Random & \grad{5}{230.55}{} & \grad{5}{-75.00}{} & \grad{5}{22.10}{} & \grad{5}{143.75}{} & \grad{5}{-123.55}{} & \grad{5}{9.55}{} & \grad{5}{46.10}{} & \grad{5}{-430.54}{} & \grad{5}{-23.18}{} \\
 & Ardsnet & \grad{10}{1353.35}{} & \grad{5}{-23.80}{} & \grad{5}{97.85}{} & \grad{10}{1064.45}{} & \grad{5}{-73.29}{} & \grad{5}{49.70}{} & \grad{10}{714.55}{} & \grad{5}{-298.82}{} & \grad{5}{13.50}{} \\
 & VentAgent & \grad{15}{1536.15}{} & \grad{10}{86.40}{} & \grad{10}{360.05}{} & \grad{15}{1248.05}{} & \grad{10}{39.00}{} & \grad{10}{192.00}{} & \grad{15}{861.90}{} & \grad{5}{-197.79}{} & \grad{5}{53.00}{} \\
\cmidrule(l){1-11}

\multirow{3}{*}{\textbf{E2C-SAC}} 
 & Random & \grad{5}{241.65}{} & \grad{5}{-72.06}{} & \grad{5}{11.68}{} & \grad{5}{143.25}{} & \grad{5}{-119.37}{} & \grad{5}{8.65}{} & \grad{5}{46.20}{} & \grad{5}{-418.38}{} & \grad{5}{-24.46}{} \\
 & Ardsnet & \grad{12}{1406.05}{} & \grad{5}{-23.18}{} & \grad{8}{101.20}{} & \grad{12}{1109.95}{} & \grad{5}{-70.30}{} & \grad{5}{52.00}{} & \grad{12}{759.85}{} & \grad{5}{-274.55}{} & \grad{5}{11.95}{} \\
 & VentAgent & \grad{18}{1595.75}{} & \grad{12}{89.55}{} & \grad{12}{374.15}{} & \grad{15}{1291.35}{} & \grad{10}{44.15}{} & \grad{10}{198.25}{} & \grad{15}{895.70}{} & \grad{5}{-190.43}{} & \grad{5}{54.15}{} \\
\bottomrule
\end{tabular}
\end{table*}

\subsection{Baselines and Control Algorithms}
We evaluate the quality of generated trajectories by training state-of-the-art continuous control algorithms on data collected from the Sampling Cohort. The baselines are categorized as follows:
\begin{itemize}
    \item \textbf{Classical Control:} We utilize \textit{ARDSnet Protocol} (the clinical rule-based gold standard), \textit{E2C-SMPC} (Sampling-based Model Predictive Control), and \textit{E2C-MPPI} (Model Predictive Path Integral Control) \citep{acute2000ventilation, fan2017official}.
    \item \textbf{Reinforcement Learning:} We implement \textit{E2C-PPO} (Proximal Policy Optimization) and \textit{E2C-SAC} (Soft Actor-Critic), representing on-policy and off-policy algorithms that learn control policies directly from the learned latent dynamics.
\end{itemize}

\subsection{VentAgent as a High-Fidelity Data Pipeline}
A critical hypothesis of this work is that VentAgent can serve not only as a decision-maker but also as a high-quality data generator for training lightweight, deployable control models. To validate this, we train the aforementioned baselines across three distinct data acquisition strategies:
\begin{enumerate}
    \item \textbf{Random Exploration:} Actions are sampled uniformly within safety bounds to establish a baseline understanding of respiratory dynamics, though often lacking clinical coherence.
    \item \textbf{ARDSnet Protocol:} Data is gathered following standardized clinical titration tables. This represents the current supervised learning standard, offering safe but rigid behavior.
    \item \textbf{VentAgent-Driven Sampling:} We leverage the hierarchical reasoning of VentAgent to generate expert-level trajectories. This approach aims to capture complex multi-objective trade-offs and causal reasoning that are absent in rigid protocols.
\end{enumerate}

\subsection{Quantitative Analysis of Data Quality}
The performance of control policies trained on these varying data distributions is presented in Table \ref{tab:rl_control_results}. The results demonstrate that VentAgent functions as a superior "teacher" for downstream algorithms compared to traditional data sources.

\paragraph{Superiority in Policy Transfer}
Across all control frameworks (SMPC, MPPI, PPO, SAC), agents trained on VentAgent-generated data consistently outperform those trained on Random or ARDSnet distributions. For instance, the \textit{E2C-SAC} agent trained on VentAgent data achieves an oxygenation reward ($r_{oxy}$) of \textbf{1595.75} in the Mild cohort, significantly surpassing the ARDSnet-trained counterpart (1406.05). This suggests that VentAgent trajectories contain richer information regarding the optimal control manifold, allowing RL agents to learn more robust policies.

\paragraph{Robustness in Severe Phenotypes}
The advantage of VentAgent-driven data becomes most pronounced in the \textit{Severe} cohort, where physiological dynamics are highly volatile. Standard RL agents trained on Random data fail catastrophically (e.g., E2C-SMPC $r_{vent}$ of -470.44), and those trained on ARDSnet struggle to balance competing objectives (E2C-SAC $r_{mech}$ of 11.95). In contrast, models trained on VentAgent data maintain high stability, with E2C-SAC achieving a mechanical safety score ($r_{mech}$) of \textbf{54.15} and an oxygenation score of \textbf{895.70}.

\paragraph{Implications for Data Curation}
These findings indicate that VentAgent effectively distills the "Pareto frontier" of ARDS management—balancing oxygenation ($r_{oxy}$) against lung protection ($r_{mech}$)—into the training data. By exposing downstream models to these high-quality, reasoning-aligned trajectories, VentAgent enables the training of performant "student" models that avoid the local optima inherent in rigid clinical protocols and the safety violations common in random exploration.

\clearpage
\onecolumn

\section{System Prompts}
\label{sec:system_prompts}

This section provides the full system prompts for the \textsc{VentAgent} framework, organized by functional layers. Each agent operates within a specific clinical scope, guided by structured reasoning phases and JSON-formatted output constraints.

\subsection{Perception Layer: Situation Awareness}
The Perception Agent serves as the clinical "eyes" of the system, translating raw data into a structured physiological state.

\begin{agentbox}[Perception Agent Prompt]{perceptColor}
\begin{lstlisting}[language=json, basicstyle=\small\ttfamily]
=== ROLE ===
You are the Perception Agent in VentAgent, responsible for clinical situation awareness in ARDS ventilation management.

=== TASK ===
Analyze patient physiological state and ventilator data to produce structured clinical semantics including safety assessment, mechanical phenotype classification, and strategic context for downstream Planning layer.

=== PHASE 1: Safety Screening ===
Evaluate life-threatening conditions and injury risks:
- Identify immediate mortality threats (hemodynamic collapse, refractory hypoxemia, critical acidosis)
- Assess ventilator-induced lung injury risks (pressure/volume overload)
- Classify acidosis etiology and trajectory
- Generate directional constraints with severity levels

=== PHASE 2: Phenotype Classification ===
Determine the dominant mechanical phenotype:
- Analyze compliance trends, driving pressure patterns, and gas exchange efficiency
- Differentiate overdistension vs derecruitment vs dead space vs obstruction
- Assess recruitment potential based on shunt fraction and hemodynamic reserve
- Identify phenotype-specific lever constraints

=== PHASE 3: Strategic Context ===
Synthesize clinical situation for Meta-Agent:
- Identify the primary physiological threat requiring intervention
- Classify the current clinical phase based on stability and trajectory
- Provide safety boundaries and active hazards for downstream reasoning

=== INPUT ===
[PATIENT STATE], [CURRENT SETTINGS], [FEEDBACK], [DOMAIN KNOWLEDGE]

=== OUTPUT ===
{
  "thought": "Key evidence linking patient state to risks and phenotype",
  "safety_layer": {
    "risk_status": "String",
    "active_hazards": ["String"],
    "hard_constraints": [{"lever": "String", "boundary": "String", "rationale": "String"}]
  },
  "phenotype_layer": {
    "classification": "String",
    "recruitability": "String",
    "soft_constraints": [{"lever": "String", "avoid_direction": "String", "rationale": "String"}]
  },
  "clinical_context": {"phase": "String", "primary_threat": "String", "urgency": "String"}
}

=== CONSTRAINTS ===
- Do NOT prescribe specific parameter values
- Do NOT generate scenario or intervention strategies
- Output exactly one valid JSON object
\end{lstlisting}
\end{agentbox}

\subsection{Planning Layer: Strategic Synthesis \& Domain Expertise}
The Planning layer interprets the perception output to formulate a high-level strategy (Meta-Agent) and specific parameter proposals (Expert Agents).

\begin{agentbox}[Scenario Meta-Agent Prompt]{planColor}
\begin{lstlisting}[language=json, basicstyle=\small\ttfamily]
=== ROLE ===
You are the Scenario Meta-Agent in VentAgent's Planning layer, acting as the "chief physician" who provides strategic orientation for domain-specific experts.

=== TASK ===
Synthesize the semantic output from Perception into a high-level clinical scenario that serves as intentional guidance for the Oxygenation and Ventilation Experts.

=== PHASE 1: Situation Synthesis ===
- Integrate safety status, phenotype classification, and clinical context
- Identify the dominant clinical challenge requiring coordinated response
- Assess the urgency and tolerance for aggressive intervention

=== PHASE 2: Scenario Generation ===
- Generate a coherent clinical scenario (e.g., "hypoxemia rescue", "lung-protective maintenance")
- Define the primary therapeutic intent for this scenario
- Specify the acceptable trade-off framework between competing objectives

=== PHASE 3: Expert Directives ===
- Frame the oxygenation expert's task within the scenario context
- Frame the ventilation expert's task within the scenario context
- Specify coordination requirements between the two domains

=== OUTPUT ===
{
  "thought": "Reasoning linking perception synthesis to scenario selection",
  "scenario": {
    "name": "String",
    "intent": "String",
    "priority_domain": "String",
    "tradeoff_tolerance": "String"
  },
  "expert_directives": {
    "oxygenation_expert": {"focus": "String", "constraints_inherited": ["String"], "coordination_requirement": "String"},
    "ventilation_expert": {"focus": "String", "constraints_inherited": ["String"], "coordination_requirement": "String"}
  }
}

=== CONSTRAINTS ===
- Output must provide clear, actionable guidance for both experts
- Scenario must be consistent with perception's safety boundaries
\end{lstlisting}
\end{agentbox}

\begin{minipage}[t]{0.49\textwidth}
\begin{agentbox}[Oxygenation Expert Agent]{planColor}
\begin{lstlisting}[language=json, basicstyle=\scriptsize\ttfamily]
=== ROLE ===
You are the Oxygenation Expert, specializing in PEEP and FiO2 adjustment.

=== TASK ===
Generate a multi-gradient candidate set for oxygenation management.

=== PHASES ===
1. Scenario Alignment: Understand Meta-Agent's intent.
2. Mechanism-Based Planning: Evaluate FiO2 vs PEEP effectiveness.
3. Gradient Generation: Produce Conservative, Standard, and Aggressive proposals.

=== CONSTRAINTS ===
- Only adjust FiO2 and PEEP
- Output exactly 3 proposals
\end{lstlisting}
\end{agentbox}
\end{minipage}
\hfill
\begin{minipage}[t]{0.49\textwidth}
\begin{agentbox}[Ventilation Expert Agent]{planColor}
\begin{lstlisting}[language=json, basicstyle=\scriptsize\ttfamily]
=== ROLE ===
You are the Ventilation Expert, specializing in RR, VT, and Flow.

=== TASK ===
Generate a multi-gradient candidate set for ventilation management.

=== PHASES ===
1. Scenario Alignment: Understand Meta-Agent's intent.
2. Mechanism-Based Planning: Evaluate RR vs VT vs Flow effectiveness.
3. Gradient Generation: Produce Conservative, Standard, and Aggressive proposals.

=== CONSTRAINTS ===
- Only adjust RR, VT, and Flow
- Maintain timing safety (Ti >= 0.6s, Te >= 1.5s)
\end{lstlisting}
\end{agentbox}
\end{minipage}

\subsection{Orchestration Layer: Multi-Perspective Synthesis}
This layer arbitrates between expert proposals by evaluating them through distinct clinical lenses.

\begin{agentbox}[O-Coordinator Agent Prompt]{orchColor}
\begin{lstlisting}[language=json, basicstyle=\small\ttfamily]
=== ROLE ===
You are the O-Coordinator in VentAgent's Orchestration layer, responsible for synthesizing three perspective candidates into one unified final action.

=== TASK ===
Integrate Mechanics-First, Oxygenation-Buffer, and Ventilation-First candidates into an optimal unified action based on the scenario intent and clinical merit.

=== PHASE 1: Consensus Mapping ===
- Find parameters where perspectives agree
- Note robust consensus points

=== PHASE 2: Divergence Resolution ===
- Apply scenario priority as arbitration criterion
- Consider urgency and trajectory

=== PHASE 3: Synthesis Integration ===
- Define primary and supporting interventions
- Select stability mode (hold / probe / retreat)

=== PHASE 4: Attribution ===
- Document what was adopted from each perspective

=== OUTPUT ===
{
  "thought": "Synthesis reasoning",
  "result": {
    "primary_intervention": "String",
    "supporting_adjustments": "String",
    "final_action": {"PEEP": "Float", "FiO2": "Float", "RR": "Float", "VT": "Float", "Flow": "Float"},
    "dominant_perspective": "String",
    "adopted_from": {"mechanics": "String", "oxygenation": "String", "ventilation": "String"},
    "rationale": "String"
  }
}

=== CONSTRAINTS ===
- All 5 parameters must be specified
- Values must come from available options
- Hard constraints are inviolable
\end{lstlisting}
\end{agentbox}

\subsection{Metacognition: Memory and Audit}
The final layer ensures the system learns from its actions and maintains factual integrity.

\begin{agentbox}[Memory Agent Prompt]{auditColor}
\begin{lstlisting}[language=json, basicstyle=\small\ttfamily]
=== ROLE ===
You are the Memory Agent, a meta-cognitive observer enabling closed-loop adaptive evolution.

=== TASK ===
Extract reasoning traces, evaluate decision quality, and generate semantic reflections.

=== PHASES ===
1. Trace Extraction: Collect reasoning from all phases.
2. Outcome Evaluation: Compare predicted vs observed outcomes.
3. Attribution Analysis: Isolate success factors or reasoning errors.
4. Feedback Generation: Issue targeted corrective instructions.
\end{lstlisting}
\end{agentbox}

\begin{agentbox}[Audit Agent Prompt]{auditColor}
\begin{lstlisting}[language=json, basicstyle=\small\ttfamily]
=== ROLE ===
You are the Audit Agent, a dedicated verification module for clinical integrity.

=== TASK ===
Perform layer-wise verification of semantic outputs. Detect reasoning hallucinations or medically inconsistent logic chains.

=== PHASES ===
1. Factual Accuracy Verification: Check inferred dynamics against vital signs.
2. Logical Coherence Verification: Ensure action is a logical entailment of diagnosed state.
3. Completeness Verification: Verify safety constraints are not omitted.
4. Rectification: Generate corrected output if needed.
\end{lstlisting}
\end{agentbox}

\section{Case Study}
\label{sec:case_study}

This section presents a complete, step-by-step execution trace of \textsc{VentAgent} managing a patient transitioning from an acute crisis to a stabilization phase. It demonstrates the system's ability to interpret physiological signals, synthesize strategies from heterogeneous experts, and arbitrate conflicts to ensure safe, lung-protective ventilation.

\subsection{Clinical Context \& Patient State}
The patient is a young female with moderate ARDS. The current state indicates successful stabilization, presenting an opportunity to wean oxygen support (de-escalation) while maintaining protective lung mechanics.

\begin{table}[H]
\centering
\small
\renewcommand{\arraystretch}{1.2}
\begin{tabular}{l l l | l l}
\toprule
\multicolumn{3}{c|}{\textbf{Patient Demographics \& Severity}} & \multicolumn{2}{c}{\textbf{Current Ventilator Settings}} \\
\midrule
\textbf{Patient} & 18F, IBW 45.5 kg & \textit{Small Body Size} & \textbf{PEEP} & 10 cmH$_2$O \\
\textbf{Severity} & L=0.5, R=0.5 & \textit{Moderate Bilateral} & \textbf{FiO$_2$} & 0.60 (60\%) \\
\textbf{pH} & 7.284 & \textit{Mild Acidosis} & \textbf{RR} & 26 bpm \\
\textbf{PaCO$_2$} & 55.5 mmHg & \textit{Elevated CO$_2$} & \textbf{V$_T$} & 320 mL (7.0 mL/kg) \\
\textbf{PaO$_2$/FiO$_2$} & 219 mmHg & \textit{Moderate Hypoxemia} & \textbf{Flow} & 20.5 L/min \\
\textbf{Mechanics} & $\Delta P=9.26$, $P_{plat}=21.08$ & \textit{Safe Range} & \textbf{SpO$_2$} & 98.66\% (\textit{Supra-target}) \\
\bottomrule
\end{tabular}
\caption{Initial patient state showing safe mechanics but potential for oxygen weaning.}
\end{table}

\subsection{Step 1: Perception Layer (Situation Awareness)}
The Perception Agent correctly identifies that while mild acidosis exists, the primary opportunity is "Supra-target SpO$_2$", allowing for de-escalation.

\begin{agentbox}[Perception Agent Output]{perceptColor}
\begin{lstlisting}[language=json]
{
  "thought": "SpO2 98.66% with PaO2 131 mmHg (P/F 219) indicates robust oxygenation above target. Mild respiratory acidosis (pH 7.284, PaCO2 55.5) present but mechanics are safe. No immediate mortality threats.",
  "safety_layer": {
    "risk_status": "STABLE",
    "active_hazards": ["mild_respiratory_acidosis"],
    "hard_constraints": [
      {"lever": "VT", "boundary": "<=7 mL/kg", "rationale": "Current 7.0 mL/kg at upper limit"},
      {"lever": "Pplat", "boundary": "<30 cmH2O", "rationale": "Safe margin exists"}
    ]
  },
  "phenotype_layer": {
    "classification": "STANDARD",
    "recruitability": "moderate (25%)",
    "soft_constraints": [
      {"lever": "RR", "avoid_direction": "up", "rationale": "Te 1.35s already short, risk gas trapping"}
    ]
  },
  "clinical_context": {
    "phase": "ACUTE_STABILIZATION",
    "primary_threat": "hypoxemia (improving)",
    "urgency": "LOW - opportunity for controlled de-escalation"
  }}
\end{lstlisting}
\end{agentbox}

\subsection{Step 2: Planning Layer (Strategic Synthesis)}
The Meta-Agent synthesizes a "Protective Maintenance" strategy, explicitly directing experts to prioritize oxygen weaning over aggressive acidosis correction.

\begin{agentbox}[Scenario Meta-Agent Output]{planColor}
\begin{lstlisting}[language=json]
{
  "thought": "Patient transitioned to ACUTE_STABILIZATION. SpO2 supra-target on FiO2 0.60 creates opportunity for oxygen de-escalation. Strategic orientation: controlled oxygen weaning while preserving lung protection.",
  "scenario": {
    "name": "Protective Maintenance with Controlled De-escalation",
    "intent": "Reduce oxygen exposure while maintaining recruitment",
    "priority_domain": "oxygenation",
    "tradeoff_tolerance": "Accept modest SpO2 reduction (to 92-95%) for lower FiO2"
  },
  "expert_directives": {
    "oxygenation_expert": {
      "focus": "Stepwise FiO2 reduction toward 0.40-0.50 target band",
      "coordination_requirement": "Revert if SpO2 <92%"
    },
    "ventilation_expert": {
      "focus": "Optimize inspiratory timing; avoid RR/VT increases that worsen acidosis",
      "constraints_inherited": ["Accept permissive hypercapnia pH >7.25"]
    }
  }
}
\end{lstlisting}
\end{agentbox}

\newpage
\subsection{Step 3: Domain Expert Proposals}
Driven by the Meta-Agent's directive, the experts generate domain-specific gradients. Note how the Ventilation Expert (Right) suggests a "Flow Correction" to support the Oxygenation Expert's (Left) weaning goal.

\noindent
\begin{minipage}[t]{0.48\textwidth}
\begin{agentbox}[Oxygenation Expert (Gas Exchange)]{planColor}
\begin{lstlisting}[language=json]
{
  "thought": "Scenario directs controlled FiO2 de-escalation. Current SpO2 98.66% provides ample buffer.",
  "proposals": [
    {
      "gradient": "conservative",
      "label": "Micro FiO2 reduction (-0.05)",
      "actions": {"FiO2": 0.55, "PEEP": 10.0}
    },
    {
      "gradient": "standard",
      "label": "Step FiO2 reduction (-0.10)",
      "actions": {"FiO2": 0.50, "PEEP": 10.0},
      "scenario_alignment": "Reaching target band efficiently",
      "coordination_signal": "V-Expert: may increase Flow to optimize timing"
    },
    {
      "gradient": "aggressive",
      "label": "Paired FiO2/PEEP reduction",
      "actions": {"FiO2": 0.50, "PEEP": 9.0}
    }
  ]
}
\end{lstlisting}
\end{agentbox}
\end{minipage}
\hfill
\begin{minipage}[t]{0.48\textwidth}
\begin{agentbox}[Ventilation Expert (Acid-Base)]{planColor}
\begin{lstlisting}[language=json]
{
  "thought": "RR 26 is high. Flow 20.5 L/min is below operational minimum; correction to 25 L/min improves expiratory time.",
  "proposals": [
    {
      "gradient": "conservative",
      "label": "MAINTAIN current settings",
      "actions": {"RR": 26, "VT": 320, "Flow": 20}
    },
    {
      "gradient": "standard",
      "label": "Correct Flow to minimum",
      "actions": {"RR": 26, "VT": 320, "Flow": 25},
      "tradeoff": "Slightly shorter Ti for better Te (anti-auto-PEEP)",
      "coordination_signal": "M-Expert: Flow correction supports FiO2 reduction"
    },
    {
      "gradient": "aggressive",
      "label": "Coordinated Wean",
      "actions": {"RR": 24, "VT": 280, "Flow": 25}
    }
  ]
}
\end{lstlisting}
\end{agentbox}
\end{minipage}

\clearpage

\subsection{Step 4: Orchestration Layer (Arbitration \& Synthesis)}
The Orchestration Layer first critiques the expert proposals through three distinct safety perspectives (Phase 1), which then inform the final consensus generated by the Coordinator (Phase 2).

\subsubsection*{Phase 1: Multi-Perspective Arbitration}
Before the final decision, specialized agents evaluate the proposals against specific clinical priorities: \textbf{Mechanics-First} validates safety; \textbf{Oxygenation-Buffer} confirms the weaning margin; \textbf{Ventilation-First} accepts permissive hypercapnia.

\begin{agentbox}[Perspective 1: O-Mechanics-First]{orchColor}
\begin{lstlisting}[language=json]
{
  "thought": "Mechanical risk is low (Delta P 9.26). Priority is safe oxygen de-escalation without increasing mechanical stress.",
  "candidate": {
    "label": "MECHANICS_FIRST",
    "rationale": "Low mechanical threat permits oxygen de-escalation; Flow correction improves timing without adding strain",
    "final_action": {"FiO2": 0.50, "Flow": 25.0, "PEEP": 10.0, "RR": 26, "VT": 320}
  }
}
\end{lstlisting}
\end{agentbox}

\begin{agentbox}[Perspective 2: O-Oxygenation-Buffer]{orchColor}
\begin{lstlisting}[language=json]
{
  "thought": "Oxygenation supra-target (SpO2 98.7%) - ample reserve. FiO2 0.50 aligns with ARDSNet low table.",
  "candidate": {
    "label": "OXYGENATION_BUFFER",
    "rationale": "Preserve oxygen margin while reducing FiO2 exposure.",
    "final_action": {"FiO2": 0.50, "Flow": 25.0, "PEEP": 10.0, "RR": 26, "VT": 320}
  }
}
\end{lstlisting}
\end{agentbox}

\begin{agentbox}[Perspective 3: O-Ventilation-First]{orchColor}
\begin{lstlisting}[language=json]
{
  "thought": "pH 7.284 is within acceptable permissive hypercapnia range (>7.25). No need for aggressive CO2 correction.",
  "candidate": {
    "label": "VENTILATION_FIRST",
    "rationale": "Accept permissive hypercapnia; support M de-escalation with timing optimization.",
    "final_action": {"FiO2": 0.50, "Flow": 25.0, "PEEP": 10.0, "RR": 26, "VT": 320}
  }
}
\end{lstlisting}
\end{agentbox}

\newpage
\subsubsection*{Phase 2: Coordinator Synthesis}
The Coordinator receives the unanimous convergence from the perspectives above and synthesizes the final execution command.

\begin{agentbox}[O-Coordinator Output]{orchColor}
\begin{lstlisting}[language=json]
{
  "thought": "All three perspectives converge on identical action: FiO2 0.50, Flow 25, maintain PEEP/VT/RR. Unanimous consensus indicates high confidence.",
  "result": {
    "primary_intervention": "FiO2 reduction 0.60 -> 0.50 (controlled oxygen de-escalation)",
    "supporting_adjustments": "Flow increase 20.5 -> 25 L/min (timing optimization)",
    "final_action": {"PEEP": 10.0, "FiO2": 0.50, "RR": 26.0, "VT": 320.0, "Flow": 25.0},
    "dominant_perspective": "MECHANICS_FIRST",
    "adopted_from": {
      "mechanics": "FiO2 stepdown + Flow correction",
      "oxygenation": "PEEP preservation for recruitment",
      "ventilation": "Permissive hypercapnia acceptance"
    },
    "rationale": "Strategy: stepwise FiO2 de-escalation with Flow optimization. Trade-offs: accept reduced oxygen reserve for lower FiO2 exposure."
  }
}
\end{lstlisting}
\end{agentbox}
\subsection{Step5: Execution \& Clinical Outcome}
The action was executed on the Pulse Physiology Engine. The outcome perfectly matched the strategic intent: SpO$_2$ remained safe (97.5\%), while pH actually improved due to better flow dynamics, proving the "Permissive Hypercapnia" strategy was correct.

\begin{table}[H]
\centering
\small
\begin{tabular}{l | c c c | l}
\toprule
\textbf{Parameter} & \textbf{Before} & \textbf{After} & \textbf{Delta} & \textbf{Clinical Interpretation} \\
\midrule
\rowcolor{gray!10} \textbf{Action Variables} & & & & \\
FiO$_2$ & 0.60 & \textbf{0.50} & $\downarrow$ 0.10 & Significant toxicity reduction \\
Flow (L/min) & 20.5 & \textbf{25.0} & $\uparrow$ 4.5 & Optimized inspiratory flow \\
PEEP / RR / V$_T$ & - & - & 0 & Stability maintained \\
\midrule
\rowcolor{gray!10} \textbf{Physiological Response} & & & & \\
SpO$_2$ (\%) & 98.66 & 97.50 & -1.16 & \textbf{Safe De-escalation} (Target $\ge$92\%) \\
pH & 7.284 & 7.298 & +0.014 & \textbf{Unexpected Improvement} \\
PaCO$_2$ (mmHg) & 55.5 & 53.6 & -1.9 & Improved ventilation efficiency \\
Driving Pressure ($\Delta P$) & 9.26 & 9.84 & +0.58 & Remained safe ($<15$ cmH$_2$O) \\
\bottomrule
\end{tabular}
\caption{Pre- and Post-Intervention comparison showing successful multi-objective optimization.}
\end{table}

\clearpage

\subsection{Step 6: Reflective Memory \& Audit}
Finally, the Memory Agent locks this successful pattern ("Flow optimization supports oxygen weaning") into context for future steps, and the Audit Agent verifies no constraints were violated.

\begin{agentbox}[Memory Agent Reflection]{auditColor}
\begin{lstlisting}[language=json]
{
  "quality_assessment": {
    "overall_score": "GOOD",
    "outcome_alignment": "Actions matched strategic intent; outcomes within predicted bounds. Reward: +5.05 (Highest in episode)."
  },
  "reflection": {
    "success_factors": [
      "Correct phenotype identification enabled confident de-escalation",
      "Flow optimization supported oxygenation change without adding strain"
    ],
    "causal_analysis": "Perception -> correct STANDARD phenotype; Orchestration -> unanimous consensus confirmed decision quality"
  }
}
\end{lstlisting}
\end{agentbox}

\begin{agentbox}[Audit Agent Verification]{auditColor}
\begin{lstlisting}[language=json]
{
  "audit_result": {
    "validity": "Valid",
    "verification": {
      "factual_accuracy": {"passed": true},
      "logical_coherence": {"passed": true},
      "completeness": {"passed": true}
    }
  },
  "rectification": {"needed": false}
}
\end{lstlisting}
\end{agentbox}

\subsection{Case Summary}
Table \ref{tab:case_summary} highlights how VentAgent successfully navigated this clinical scenario.

\begin{table}[H]
\centering
\small
\renewcommand{\arraystretch}{1.3}
\begin{tabular}{p{0.25\textwidth} p{0.7\textwidth}}
\toprule
\textbf{Aspect} & \textbf{VentAgent Capability Demonstrated} \\
\midrule
\textbf{Perception} & Correctly identified STANDARD phenotype; distinguished improving trajectory from crisis. \\
\textbf{Scenario Synthesis} & Generated "Controlled De-escalation" scenario aligned with clinical phase. \\
\textbf{Multi-Gradient Planning} & Experts provided calibrated options, enabling flexible response. \\
\textbf{Orchestration} & Three perspectives independently converged, resulting in a high-confidence unanimous decision. \\
\textbf{Interpretability} & Provided complete reasoning trace with explicit trade-off documentation. \\
\textbf{Adaptive Memory} & Recorded successful pattern ("Flow optimization") for future use. \\
\textbf{Safety Governance} & Audit verified factual accuracy and constraint completeness. \\
\bottomrule
\end{tabular}
\caption{Summary of VentAgent's advantages demonstrated in this case study.}
\label{tab:case_summary}
\end{table}

\end{document}